\documentclass{article}

\PassOptionsToPackage{numbers, compress}{natbib}
\usepackage{neurips_data_2024}

\usepackage[utf8]{inputenc} 
\usepackage[T1]{fontenc}    
\usepackage{hyperref}       
\usepackage{url}            
\usepackage{booktabs}       
\usepackage{amsfonts}       
\usepackage{nicefrac}       
\usepackage{microtype}      
\usepackage{xcolor}         
\usepackage{graphicx} 
\usepackage{makecell}
\usepackage{multirow} 
\usepackage{colortbl}
\usepackage{pifont}
\usepackage{authblk}

\usepackage{titletoc}
\usepackage{longtable}
\usepackage{listings}
\usepackage{tcolorbox}
\usepackage{longtable}
\usepackage{tabularx} 
\usepackage{amsmath}
\usepackage{xcolor}

\usepackage{color}
\usepackage{soul}
\definecolor{lightred}{rgb}{1, 0.8, 0.8}
\definecolor{lightblue}{rgb}{0.8, 0.9, 1}

\newcommand{\greencheck}{\textcolor{green}{\ding{52}}}%
\newcommand{\xmark}{\textcolor{red}{\ding{55}}}%

\title{\textsc{MedCalc-Bench}: Evaluating Large Language Models for Medical Calculations}

\author[1]{Nikhil Khandekar$^*$}
\author[1]{Qiao Jin$^*$}
\author[2]{Guangzhi Xiong$^*$}
\author[3, 4]{Soren Dunn}
\author[5]{Serina S Applebaum}
\author[7]{Zain Anwar}
\author[8]{Maame Sarfo-Gyamfi}
\author[5]{Conrad W Safranek}
\author[6]{Abid A Anwar}
\author[9]{Andrew Zhang}
\author[5]{Aidan Gilson}
\author[5]{Maxwell B Singer}
\author[5]{Amisha Dave}
\author[5]{Andrew Taylor}
\author[2]{\\Aidong Zhang}
\author[5]{Qingyu Chen}
\author[1]{Zhiyong Lu$^\dag$}

\makeatletter
\renewcommand\AB@affilsepx{, \protect\Affilfont} 
\makeatother

\affil[1]{National Library of Medicine, National Institutes of Health}
\affil[2]{University of Virginia}
\affil[3]{University of Illinois at Urbana Champaign}
\affil[4]{Lapis Labs}
\affil[5]{Yale University School of Medicine}
\affil[6]{University of Illinois College of Medicine at Chicago}
\affil[7]{Rosalind Franklin University Chicago Medical School}
\affil[8]{Howard University College of Medicine}
\affil[9]{University of Chicago Pritzker School of Medicine}

\begin{document}

\maketitle

\def\thefootnote{$*$}\footnotetext{Equal contribution. $^\dagger$Correspondence: \nolinkurl{zhiyong.lu@nih.gov}.}

\renewcommand{\thefootnote}{\arabic{footnote}} 
\setcounter{footnote}{0} 

\begin{abstract}
As opposed to evaluating computation and logic-based reasoning, current benchmarks for evaluating large language models (LLMs) in medicine are primarily focused on question-answering involving domain knowledge and descriptive reasoning. While such qualitative capabilities are vital to medical diagnosis, in real-world scenarios, doctors frequently use clinical calculators that follow quantitative equations and rule-based reasoning paradigms for evidence-based decision support. To this end, we propose \textsc{MedCalc-Bench}, a first-of-its-kind dataset focused on evaluating the medical calculation capability of LLMs. \textsc{MedCalc-Bench} contains an evaluation set of over 1000 manually reviewed instances from 55 different medical calculation tasks. Each instance in \textsc{MedCalc-Bench} consists of a patient note, a question requesting to compute a specific medical value, a ground truth answer, and a step-by-step explanation showing how the answer is obtained. While our evaluation results show the potential of LLMs in this area, none of them are effective enough for clinical settings. Common issues include extracting the incorrect entities, not using the correct equation or rules for a calculation task, or incorrectly performing the arithmetic for the computation. We hope our study highlights the quantitative knowledge and reasoning gaps in LLMs within medical settings, encouraging future improvements of LLMs for various clinical calculation tasks. \footnote{\textsc{MedCalc-Bench} is publicly available at: \url{https://github.com/ncbi-nlp/MedCalc-Bench}.}
\end{abstract}

\section{Introduction}

\label{intro}
Large language models (LLMs) such as GPT \cite{brown2020language, openai2023gpt4}, Gemini/PaLM \cite{anil2023palm, team2023gemini}, and Llama \cite{touvron2023llama1, touvron2023llama2} have been successfully applied to a variety of biomedical tasks \cite{nori2023capabilities, saab2024capabilities, thirunavukarasu2023large, tian2024opportunities}, including but not limited to question answering \cite{lievin2024can, nori2023can, singhal2023large}, clinical trial matching \cite{jin2023matching, wong2023scaling, wornow2024zero, zhuang2024team}, and medical document summarization \cite{shaib2023summarizing, tang2023evaluating, van2024adapted}.
However, most of these tasks have a limited evaluation of domain knowledge and qualitative reasoning ability of LLMs, as demonstrated by the commonly used medical benchmarks such as MedQA \cite{jin2021disease}, PubMedQA \cite{jin2019pubmedqa}, and MedMCQA \cite{pal2022medmcqa}.
While quantitative tools such as medical calculators are frequently used in clinical settings \cite{dziadzko2016clinical, green2019medical}, currently there is no benchmark evaluating the medical calculation capabilities of LLMs.

\begin{figure}
    \centering
    \includegraphics[width=1\linewidth]{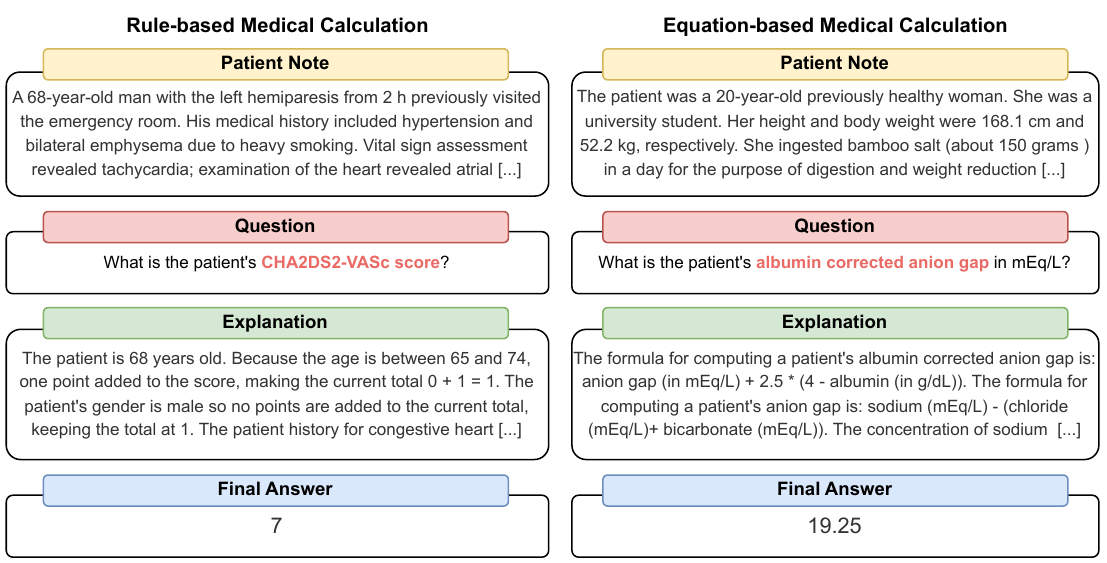}
    \caption{Example instances of the \textsc{MedCalc-Bench} dataset. }
    \label{fig:examples}
\end{figure}

Medical calculators are statistical tools derived from high-quality clinical studies, serving various purposes, including metric conversions \cite{cockcroft1976prediction}, disease diagnosis \cite{initiative20102010}, and prognosis prediction \cite{gage2001validation}.
Figure \ref{fig:examples} shows two examples of medical calculators.
To accurately compute requested medical scores, the model needs to have three non-trivial capabilities: (1) possessing the knowledge of the rules or equations for the medical calculation task, (2) identifying and extracting the relevant parameters within a long patient note, and (3) conducting the arithmetic computation for the task correctly. 

In this work, we propose \textsc{MedCalc-Bench}, a first-of-its-kind dataset for evaluating the medical calculation capabilities of LLMs.
To construct \textsc{MedCalc-Bench}, we first curated 55 common medical calculation tasks from MDCalc\footnote{\url{https://www.mdcalc.com/\#Popular}}. 
Then, we compiled Open-Patients, a collection of over 180k publicly available patient notes, and identified the notes that can be used for each calculation task.
Finally, we collected over 1k instances for \textsc{MedCalc-Bench}, where each instance contains: (1) a patient note, (2) a question requesting to compute a specific medical value, (3) a manually reviewed ground truth answer, and (4) a step-by-step explanation of the computation process. 

Using \textsc{MedCalc-Bench}, we conducted systematic evaluations of various LLMs, including the state-of-the-art proprietary models such as GPT-4 \cite{openai2023gpt4}, open-source LLMs such as Llama \cite{touvron2023llama2} and Mixtral \cite{jiang2024mixtral}, as well as biomedical domain-specific PMC-LLaMA \cite{wu2024pmc} and \textsc{MediTron} \cite{chen2023meditron}.
Our experimental results show that most of the tested models struggle in the task. 
GPT-4 achieved the best baseline performance of only 50.9\% accuracy using one-shot chain-of-thought prompting.
By analyzing the types of errors made by LLMs, we found that the models suffer mostly from insufficient medical calculator knowledge in the zero-shot setting. To mitigate this issue, we add a one-shot exemplar in the prompt, showing the model how to apply the requested medical equations or rules. 
Our analysis revealed additional issues in extracting calculator-related attributes and in arithmetic computations.
These results can provide insights into future improvement in the medical calculation capabilities of LLMs.

In summary, the contributions of our study are threefold:
\begin{itemize}
    \item We manually curated \textsc{MedCalc-Bench}, a novel dataset of over 1k instances for evaluating the capabilities of LLMs across 55 different medical calculation tasks.
    \item We conducted comprehensive evaluations on \textsc{MedCalc-Bench} with various open and closed-source LLMs. Our results show that all current LLMs are not yet ready for medical calculations, with the best accuracy of only 50.9\% achieved by GPT-4.
    \item Our error analysis reveals the insufficiency of calculator knowledge in LLMs, as well as their deficiencies in attribute extraction and arithmetic computation for medical calculation.
\end{itemize}

\section{\textsc{MedCalc-Bench}}

\label{headings}

\subsection{Calculation Task Curation}

\label{sec:task_curation}
\textsc We selected 55 different calculators for \textsc{MedCalc-Bench}, all of which were listed as ``popular'' on MDCalc, the most commonly used online medical calculator website by clinicians \cite{elovic2019mdcalc}.
As shown in Figure \ref{fig:examples}, they fall into two major categories, \textbf{rule-based} calculation (19 calculators) and  \textbf{equation-based} calculation (36 calculators). 

Rule-based calculators typically contain a list of criteria, where each criterion is a condition of a specific medical attribute.
An instance of this would be the HEART score calculator \cite{six2008chest}, which takes in both numerical attributes such as the patient's age (e.g., if the patient is older than 65 years, add two points; if the patient's age is between 45 and 64, add one point; and zero points otherwise) and categorical variables such as the presence of significant ST elevation (adding two points if present; zero points otherwise). The final answer for these calculators will be a discrete answer after taking the sum of the sub-scores. 

Like rule-based calculators, equation-based calculators also take in both categorical (e.g., gender, race) and numerical variables (e.g., creatinine concentration, age, and height). However, equation-based calculators follow a specific formula to output a decimal, date, or time given the attributes instead of additively combining sub-scores for each criterion. An instance of an equation-based calculator would be the MDRD GRF equation \cite{levey1999more}. This equation computes the patient's eGFR, using the patient's gender and race as coefficients in addition to the patient's creatinine concentration. The only equation-based calculators which do not output a decimal are Estimated Due Date (EDD), Estimated Date of Conception (EDC), and Estimated Gestational Age (EGA). These three calculators compute a date (for EDC, EGA) or a time (for EGA) instead. 

\textsc For each instance, \textsc{MedCalc-Bench} also provides a natural language explanation for how the final answer is computed.
We implement template-based explanation generators for each of the 55 calculators. These templates first list the numerical and categorical variable values, and then plug them in to show how the final answers are obtained. 
The implementation details can be found in supplementary materials. 

\subsection{Dataset Instance Collection}

\label{sec:collection}

In this section, we describe how patient notes and answers were collected for the 55 different calculation tasks in \textsc{MedCalc-Bench}. We aimed to collect at most 20 notes for each calculator. Specifically, the patient notes were collected using the following three-step pipeline. 

\textbf{(1) Note collection and attribute extraction.} We compiled Open-Patients\footnote{Publicly available at \url{https://huggingface.co/datasets/ncbi/Open-Patients}.}, a collection of over 180k public patient notes, including anonymized real case reports from PMC-Patients \cite{zhao2023large}, case vignettes in MedQA-USMLE \cite{jin2021disease}, synthetic cases in TREC Clinical Decision Support Tracks \cite{simpson2014overview, roberts2015overview} and TREC Clinical Trials Tracks \cite{roberts2021overview}.
Using GPT-3.5-Turbo, we identified patient notes for each calculator based on its eligibility criteria. We then used GPT-4 to extract the attribute values needed for each calculator from the eligible notes. 

\textbf{(2) Data verification and enrichment.} The extracted values were verified and corrected by medical professionals. After the verification, 34 of the 55 calculators have at least one eligible note with the extracted attribute required for computation. Of the remaining 21 calculators, 10 of them are equation-based calculators, for which we generated 20 synthesized notes for each of them using corresponding templates. The other 11 calculators are rule-based for which clinicians synthesized 20 patient notes each. 

\textbf{(3) Answer and explanation generation.} 
After obtaining patient notes with the extracted values, for each of the 55 calculators, we generated step-by-step explanations to derive the final answers. Specifically, we implemented templates for each calculator to generate the natural language explanations.
From these three steps, we curated 1047 instances for \textsc{MedCalc-Bench}, each of which contains a patient note, a question, along with a ground-truth explanation and final answer. 

\subsection{Dataset Characteristics}

Table~\ref{tab:dataset_statistics} shows the statistics of \textsc{MedCalc-Bench} and the different calculator sub-types. The equation-based calculators have between 1 to 7 attributes, while the rule-based calculators have between 3 to 31 attributes. Thus, it may require a varying number of reasoning steps to solve different tasks in our dataset. 

\begin{table}[h!] \small
    \centering
    \caption{Statistics of \textsc{MedCalc-Bench}. Inst.: instance; Avg.: average; Attr.: attribute; Q.: question.}
    \begin{tabular}{lcccccccc}
    \toprule
     & \bf \makecell{\#Tasks} & \bf \makecell{\#Inst.} & \bf \makecell{Avg. L\\of Note} & \bf \makecell{Avg. L\\of Q.} & \bf \makecell{Min\\Attr.} & \bf \makecell{Max\\Attr.} & \bf \makecell{Avg.\\Attr.} & \bf \makecell{Example Calculation} \\
    \midrule
    \multicolumn{9}{l}{\textbf{Equation-based Calculation Tasks}}\\
    \midrule
    \includegraphics[width=0.4cm]{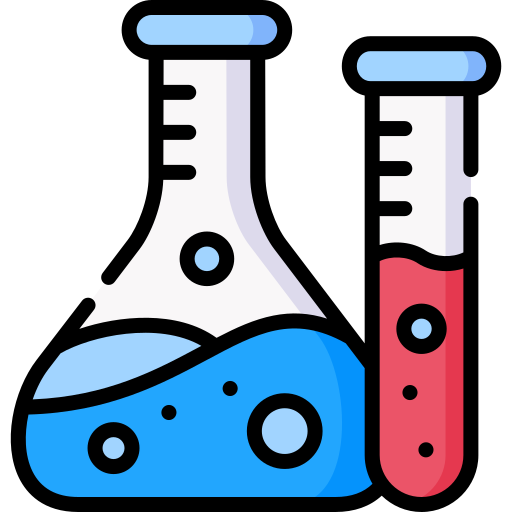} Lab & 19 & 327 & 891.0 & 22.3 & 2 & 7 & 3.6 & LDL Concentration \\
    \includegraphics[width=0.4cm]{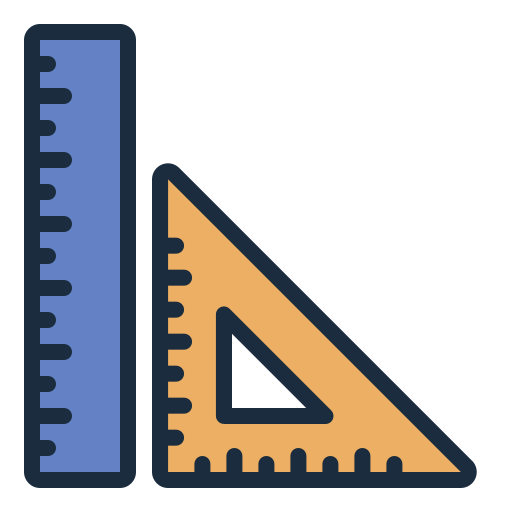} Physical & 12 & 240 & 419.3 & 20.8 & 1 & 3 & 2.0 & QTc (Bazett Formula) \\
    \includegraphics[width=0.4cm]{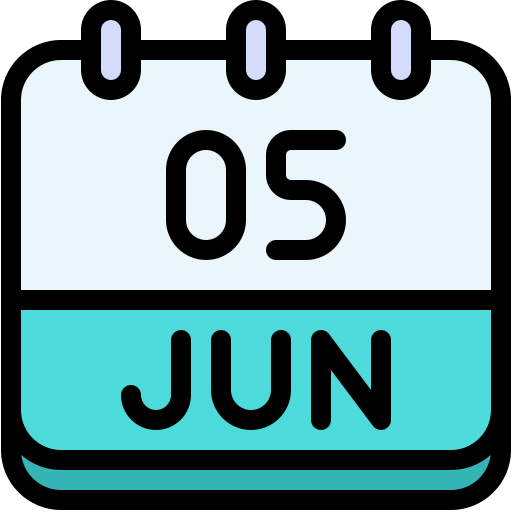} Date & 3 & 60 & 25.3 & 67.0 & 2 & 2 & 2.0 & Estimated Due Date \\
    \includegraphics[width=0.4cm]{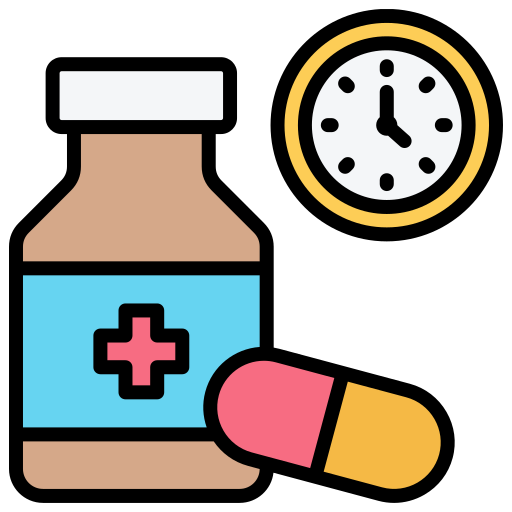} Dosage & 2 & 40 & 31.4 & 31.0 & 2 & 6 & 4.0 & Morphine Equivalents \\
    \midrule
    \multicolumn{9}{l}{\textbf{Rule-based Calculation Tasks}}\\
    \midrule
    \includegraphics[width=0.4cm]{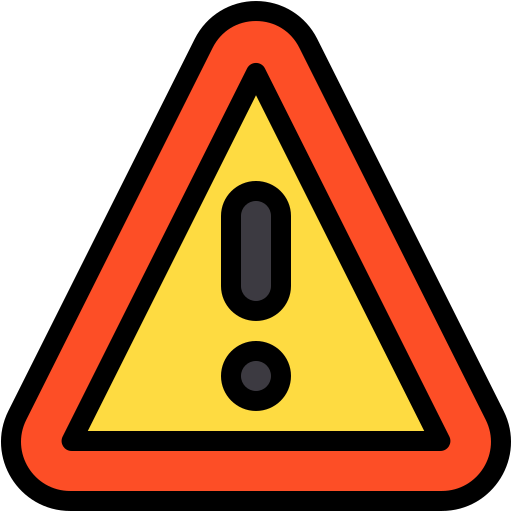} Risk & 12 & 240 & 422.1 & 14.9 & 5 & 31 & 11.5 & Caprini Score for VTE \\
    \includegraphics[width=0.4cm]{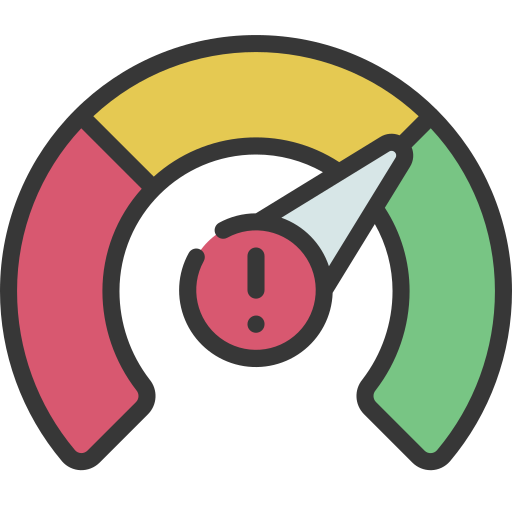} Severity & 4 & 80 & 262.6 & 11.0 & 3 & 20 & 7.7 & Pneumonia Severity Idx \\
    \includegraphics[width=0.4cm]{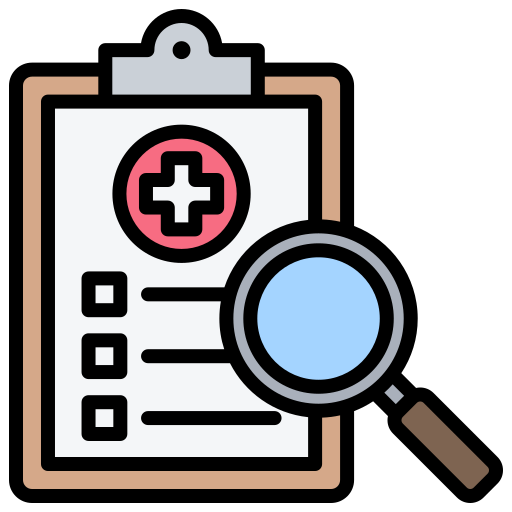} Diagnosis & 3 & 60 & 625.6 & 15.0 & 3 & 9 & 5.3 & PERC Rule for PE \\
    \midrule
    \textbf{Overall} & 55 & 1047 & 529.7 & 21.9 & 1 & 31 & 5.4 & -- \\
    \bottomrule
    \end{tabular}
    \label{tab:dataset_statistics}
\end{table}

Our dataset evaluates three distinct  capabilities required for medical calculation:

\textbf{(1) Recall of medical calculation knowledge.}
The first required capability is the recall of correct medical knowledge for given questions from seven different domains shown in Table \ref{tab:dataset_statistics}. As mentioned above, medical calculators can have various sub-types, which challenge LLMs to recall the exact knowledge, including medical equations or rules, to solve the clinical calculation task. 

\textbf{(2) Extraction of relevant patient attributes.}
The second required capability is the extraction of correct attributes from patient notes, given the noises in the long context of over 500 words on average. LLMs are required to extract both numerical and categorical attributes. The medical context complicates such extractions, with the existence of multiple synonyms (e.g., both HbA1c and glycohemoglobin denote the same entity) and the requirement of determining the presence of attributes (e.g., a blood pressure of 160/100 mmHg indicates the presence of hypertension) using medical knowledge and clinical reasoning.

\textbf{(3) Arithmetic computation of the final results.}
The third required capability is the computation of final results, especially the derivation of scores through multi-step reasoning. While datasets like GSM-8k \cite{cobbe2021training} have tested the arithmetic calculation capability of LLMs, \textsc{MedCalc-Bench} presents a more challenging task, as it requires LLMs to fully understand the sequence and dependencies among multiple medical equations or rules, in order to make the correct computation.
Additionally, \textsc{MedCalc-Bench} also contains some exponential computations that are not covered by other math datasets.

Overall, we believe that \textsc{MedCalc-Bench} serves as a comprehensive benchmark which not only examines the internal medical calculation knowledge of LLMs, but also tests general-purpose skills such as attribute extraction and arithmetic computation in a more challenging domain-specific setting. 

\section{Evaluation}

\subsection{Settings}

\label{sec:settings}
To establish the baseline performance in \textsc{MedCalc-Bench}, we experiment with eight different LLMs under three common prompting strategies.
Specifically, three groups of LLMs are considered: 
\textbf{Medical domain-specific} LLMs include PMC-LLaMA-13B \cite{wu2024pmc} and \textsc{MediTron}-70B \cite{chen2023meditron}; 
\textbf{Proprietary} LLMs include GPT-4 \cite{openai2023gpt4} and GPT-3.5 \cite{ouyang2022training}; \textbf{Open-source} LLMs, including 8B and 70B Llama 3 \cite{touvron2023llama2}, as well as Mistral-7B \cite{jiang2023mistral} and Mixtral-8x7B \cite{jiang2024mixtral}.

Similarly, we consider three prompting strategies: 
\textbf{Zero-shot Direct Prompting}: In this setting, the LLM is prompted to directly output the answer without any explanation;
\textbf{Zero-shot Chain-of-Thought (CoT) Prompting}: In this setting, the LLM is prompted to first generate step-by-step rationale and then generate the answer \cite{wei2022chain};
\textbf{One-shot CoT Prompting}: In this setting, the LLM is provided with an exemplar of the corresponding calculation task.
The exemplar is manually curated and contains the patient note, question, and the output consisting of the step-by-step explanation and final answer value. 

Based on the output type, we have three different evaluation settings: (1) For all rule-based calculators, the final answer must be the exact same as the ground-truth answer, (2) For equation-based calculators that are lab tests, physical tests, and dosage conversion calculators, the predicted answer must be within 5\% of the ground-truth answer, (3) For date-based equation calculations, the predicted dates should exactly match the ground truth. 

\begin{table}[h!] \small
    \centering
    \caption{\textsc{MedCalc-Bench} accuracy of different systems. All numbers are in percentages. Phys.: Physical; Sev.: Severity; Diag.: Diagnosis; Avg.: Average; $\pm$ std for all results are shown.}
    \begin{tabular}{lc|cccc|ccc|c}
        \toprule
        \multirow{3}{*}{\bf Model} & \multirow{3}{*}{\bf Size} & \multicolumn{4}{c|}{\bf Equation} & \multicolumn{3}{c|}{\bf Rule-based} & \multirow{3}{*}{\bf Avg.} \\
        \cmidrule(lr){3-9}
         &  & Lab & Phys. & Date & Dosage & Risk & Sev. & Diag. & \\
        \midrule
        \rowcolor[RGB]{234, 238, 234}
        \multicolumn{10}{c}{\bf Zero-shot Direct Prompting} \\
        \midrule
\includegraphics[width=0.4cm]{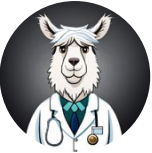} PMC-LLaMA \cite{wu2024pmc} & 13B & \makecell{0.00\\\textcolor{gray}{\scriptsize$\pm$0.00}}  & \makecell{0.00\\\textcolor{gray}{\scriptsize$\pm$0.00}}  & \makecell{0.00\\\textcolor{gray}{\scriptsize$\pm$0.00}}  & \makecell{0.00\\\textcolor{gray}{\scriptsize$\pm$0.00}}  & \makecell{0.00\\\textcolor{gray}{\scriptsize$\pm$0.00}}  & \makecell{0.00\\\textcolor{gray}{\scriptsize$\pm$0.00}}  & \makecell{0.00\\\textcolor{gray}{\scriptsize$\pm$0.00}}  & \makecell{0.00\\\textcolor{gray}{\scriptsize$\pm$0.00}}  \\
\includegraphics[width=0.4cm]{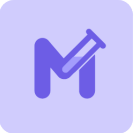} \textsc{MediTron} \cite{chen2023meditron} & 70B & \makecell{3.67\\\textcolor{gray}{\scriptsize$\pm$0.01}}  & \makecell{8.33\\\textcolor{gray}{\scriptsize$\pm$0.02}}  & \makecell{5.00\\\textcolor{gray}{\scriptsize$\pm$0.03}}  & \makecell{0.00\\\textcolor{gray}{\scriptsize$\pm$0.00}}  & \makecell{7.50\\\textcolor{gray}{\scriptsize$\pm$0.02}}  & \makecell{5.00\\\textcolor{gray}{\scriptsize$\pm$0.02}}  & \makecell{13.33\\\textcolor{gray}{\scriptsize$\pm$0.04}}  & \makecell{6.21\\\textcolor{gray}{\scriptsize$\pm$0.01}}  \\
\includegraphics[width=0.4cm]{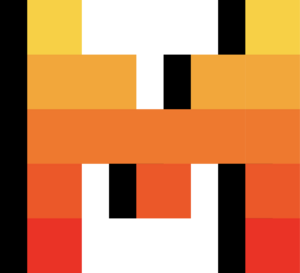} Mistral \cite{jiang2023mistral} & 7B & \makecell{10.70\\\textcolor{gray}{\scriptsize$\pm$0.02}}  & \makecell{18.33\\\textcolor{gray}{\scriptsize$\pm$0.02}}  & \makecell{3.33\\\textcolor{gray}{\scriptsize$\pm$0.02}}  & \makecell{0.00\\\textcolor{gray}{\scriptsize$\pm$0.00}}  & \makecell{4.58\\\textcolor{gray}{\scriptsize$\pm$0.01}}  & \makecell{3.75\\\textcolor{gray}{\scriptsize$\pm$0.02}}  & \makecell{13.33\\\textcolor{gray}{\scriptsize$\pm$0.04}}  & \makecell{9.84\\\textcolor{gray}{\scriptsize$\pm$0.01}}  \\
\includegraphics[width=0.4cm]{icons/mistral.png} Mixtral \cite{jiang2024mixtral} & 8x7B & \makecell{12.23\\\textcolor{gray}{\scriptsize$\pm$0.02}}  & \makecell{23.33\\\textcolor{gray}{\scriptsize$\pm$0.03}}  & \makecell{5.00\\\textcolor{gray}{\scriptsize$\pm$0.03}}  & \makecell{7.50\\\textcolor{gray}{\scriptsize$\pm$0.04}}  & \makecell{12.92\\\textcolor{gray}{\scriptsize$\pm$0.02}}  & \makecell{7.50\\\textcolor{gray}{\scriptsize$\pm$0.03}}  & \makecell{16.67\\\textcolor{gray}{\scriptsize$\pm$0.05}}  & \makecell{14.23\\\textcolor{gray}{\scriptsize$\pm$0.01}}  \\
\includegraphics[width=0.4cm]{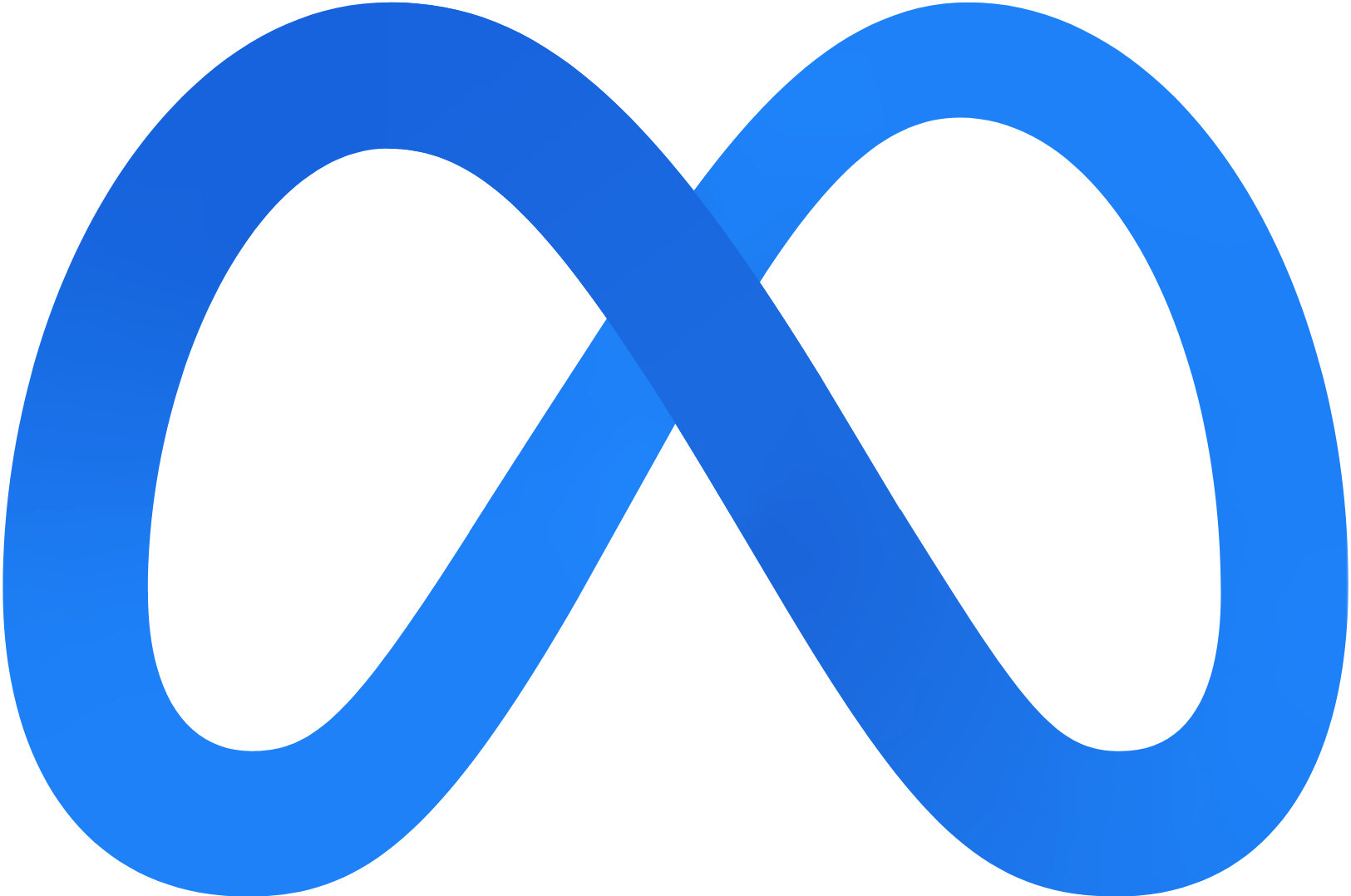} Llama 3 \cite{touvron2023llama2} & 8B & \makecell{10.70\\\textcolor{gray}{\scriptsize$\pm$0.02}}  & \makecell{19.17\\\textcolor{gray}{\scriptsize$\pm$0.03}}  & \makecell{3.33\\\textcolor{gray}{\scriptsize$\pm$0.02}}  & \makecell{5.00\\\textcolor{gray}{\scriptsize$\pm$0.03}}  & \makecell{12.50\\\textcolor{gray}{\scriptsize$\pm$0.02}}  & \makecell{8.75\\\textcolor{gray}{\scriptsize$\pm$0.03}}  & \makecell{25.00\\\textcolor{gray}{\scriptsize$\pm$0.06}}  & \makecell{13.09\\\textcolor{gray}{\scriptsize$\pm$0.01}}  \\
\includegraphics[width=0.4cm]{icons/meta.png} Llama 3 \cite{touvron2023llama2} & 70B & \makecell{18.04\\\textcolor{gray}{\scriptsize$\pm$0.02}}  & \makecell{33.33\\\textcolor{gray}{\scriptsize$\pm$0.03}}  & \makecell{8.33\\\textcolor{gray}{\scriptsize$\pm$0.04}}  & \makecell{12.50\\\textcolor{gray}{\scriptsize$\pm$0.05}}  & \makecell{15.83\\\textcolor{gray}{\scriptsize$\pm$0.02}}  & \makecell{13.75\\\textcolor{gray}{\scriptsize$\pm$0.04}}  & \makecell{33.33\\\textcolor{gray}{\scriptsize$\pm$0.06}}  & \makecell{20.82\\\textcolor{gray}{\scriptsize$\pm$0.01}}  \\
\includegraphics[width=0.4cm]{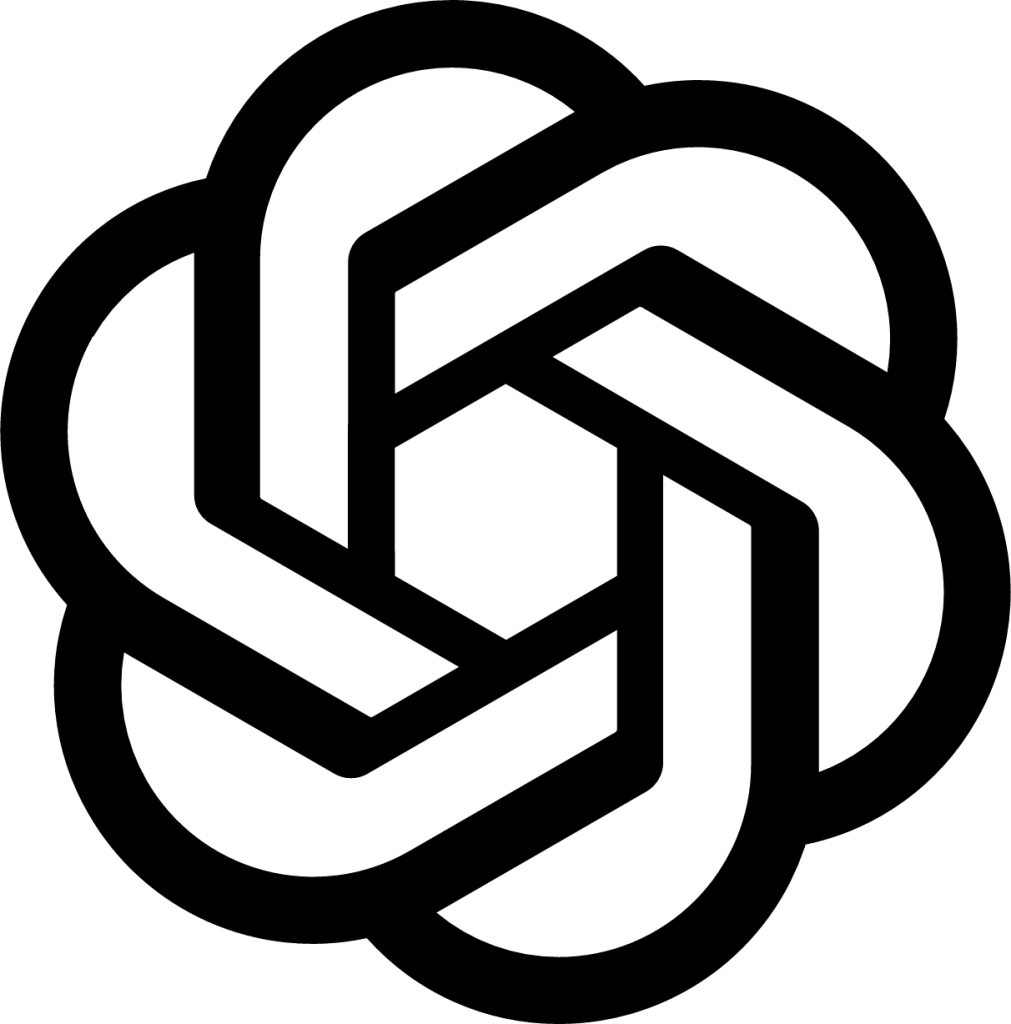} GPT-3.5 \cite{ouyang2022training} & N/A & \makecell{17.13\\\textcolor{gray}{\scriptsize$\pm$0.02}}  & \makecell{35.00\\\textcolor{gray}{\scriptsize$\pm$0.03}}  & \makecell{13.33\\\textcolor{gray}{\scriptsize$\pm$0.04}}  & \makecell{5.00\\\textcolor{gray}{\scriptsize$\pm$0.03}}  & \makecell{12.92\\\textcolor{gray}{\scriptsize$\pm$0.02}}  & \makecell{6.25\\\textcolor{gray}{\scriptsize$\pm$0.03}}  & \makecell{18.33\\\textcolor{gray}{\scriptsize$\pm$0.05}}  & \makecell{18.82\\\textcolor{gray}{\scriptsize$\pm$0.01}}  \\
\includegraphics[width=0.4cm]{icons/openai.png} GPT-4 \cite{openai2023gpt4} & N/A & \makecell{14.37\\\textcolor{gray}{\scriptsize$\pm$0.02}}  & \makecell{34.58\\\textcolor{gray}{\scriptsize$\pm$0.03}}  & \makecell{38.33\\\textcolor{gray}{\scriptsize$\pm$0.06}}  & \makecell{15.00\\\textcolor{gray}{\scriptsize$\pm$0.06}}  & \makecell{14.58\\\textcolor{gray}{\scriptsize$\pm$0.02}}  & \makecell{15.00\\\textcolor{gray}{\scriptsize$\pm$0.04}}  & \makecell{20.00\\\textcolor{gray}{\scriptsize$\pm$0.05}}  & \makecell{20.82\\\textcolor{gray}{\scriptsize$\pm$0.01}}  \\
        \midrule
        \rowcolor[RGB]{234, 238, 234}
        \multicolumn{10}{c}{\bf Zero-shot CoT Prompting} \\
        \midrule
\includegraphics[width=0.4cm]{icons/pmc.png} PMC-LLaMA \cite{wu2024pmc} & 13B & \makecell{0.00\\\textcolor{gray}{\scriptsize$\pm$0.00}}  & \makecell{0.00\\\textcolor{gray}{\scriptsize$\pm$0.00}}  & \makecell{0.00\\\textcolor{gray}{\scriptsize$\pm$0.00}}  & \makecell{0.00\\\textcolor{gray}{\scriptsize$\pm$0.00}}  & \makecell{0.00\\\textcolor{gray}{\scriptsize$\pm$0.00}}  & \makecell{0.00\\\textcolor{gray}{\scriptsize$\pm$0.00}}  & \makecell{0.00\\\textcolor{gray}{\scriptsize$\pm$0.00}}  & \makecell{0.00\\\textcolor{gray}{\scriptsize$\pm$0.00}}  \\
\includegraphics[width=0.4cm]{icons/meditron.png} \textsc{MediTron} \cite{chen2023meditron} & 70B & \makecell{0.00\\\textcolor{gray}{\scriptsize$\pm$0.00}}  & \makecell{0.00\\\textcolor{gray}{\scriptsize$\pm$0.00}}  & \makecell{3.33\\\textcolor{gray}{\scriptsize$\pm$0.02}}  & \makecell{0.00\\\textcolor{gray}{\scriptsize$\pm$0.00}}  & \makecell{0.00\\\textcolor{gray}{\scriptsize$\pm$0.00}}  & \makecell{0.00\\\textcolor{gray}{\scriptsize$\pm$0.00}}  & \makecell{3.33\\\textcolor{gray}{\scriptsize$\pm$0.02}}  & \makecell{0.38\\\textcolor{gray}{\scriptsize$\pm$0.00}}  \\
\includegraphics[width=0.4cm]{icons/mistral.png} Mistral \cite{jiang2023mistral} & 7B & \makecell{10.09\\\textcolor{gray}{\scriptsize$\pm$0.02}}  & \makecell{14.58\\\textcolor{gray}{\scriptsize$\pm$0.02}}  & \makecell{1.67\\\textcolor{gray}{\scriptsize$\pm$0.02}}  & \makecell{0.00\\\textcolor{gray}{\scriptsize$\pm$0.00}}  & \makecell{9.58\\\textcolor{gray}{\scriptsize$\pm$0.02}}  & \makecell{7.50\\\textcolor{gray}{\scriptsize$\pm$0.03}}  & \makecell{25.00\\\textcolor{gray}{\scriptsize$\pm$0.06}}  & \makecell{10.79\\\textcolor{gray}{\scriptsize$\pm$0.01}}  \\
\includegraphics[width=0.4cm]{icons/mistral.png} Mixtral \cite{jiang2024mixtral} & 8x7B & \makecell{22.63\\\textcolor{gray}{\scriptsize$\pm$0.02}}  & \makecell{40.00\\\textcolor{gray}{\scriptsize$\pm$0.03}}  & \makecell{6.67\\\textcolor{gray}{\scriptsize$\pm$0.03}}  & \makecell{17.50\\\textcolor{gray}{\scriptsize$\pm$0.06}}  & \makecell{11.25\\\textcolor{gray}{\scriptsize$\pm$0.02}}  & \makecell{21.25\\\textcolor{gray}{\scriptsize$\pm$0.05}}  & \makecell{15.00\\\textcolor{gray}{\scriptsize$\pm$0.05}}  & \makecell{22.35\\\textcolor{gray}{\scriptsize$\pm$0.01}}  \\
\includegraphics[width=0.4cm]{icons/meta.png} Llama 3 \cite{touvron2023llama2} & 8B & \makecell{16.51\\\textcolor{gray}{\scriptsize$\pm$0.02}}  & \makecell{25.00\\\textcolor{gray}{\scriptsize$\pm$0.03}}  & \makecell{1.67\\\textcolor{gray}{\scriptsize$\pm$0.02}}  & \makecell{7.50\\\textcolor{gray}{\scriptsize$\pm$0.04}}  & \makecell{11.25\\\textcolor{gray}{\scriptsize$\pm$0.02}}  & \makecell{13.75\\\textcolor{gray}{\scriptsize$\pm$0.04}}  & \makecell{26.67\\\textcolor{gray}{\scriptsize$\pm$0.06}}  & \makecell{16.43\\\textcolor{gray}{\scriptsize$\pm$0.01}}  \\
\includegraphics[width=0.4cm]{icons/meta.png} Llama 3 \cite{touvron2023llama2} & 70B & \makecell{33.94\\\textcolor{gray}{\scriptsize$\pm$0.03}}  & \makecell{66.25\\\textcolor{gray}{\scriptsize$\pm$0.03}}  & \makecell{25.00\\\textcolor{gray}{\scriptsize$\pm$0.06}}  & \makecell{20.00\\\textcolor{gray}{\scriptsize$\pm$0.06}}  & \makecell{18.33\\\textcolor{gray}{\scriptsize$\pm$0.02}}  & \makecell{16.25\\\textcolor{gray}{\scriptsize$\pm$0.04}}  & \makecell{36.67\\\textcolor{gray}{\scriptsize$\pm$0.06}}  & \makecell{35.53\\\textcolor{gray}{\scriptsize$\pm$0.01}}  \\
\includegraphics[width=0.4cm]{icons/openai.png} GPT-3.5 \cite{ouyang2022training} & N/A & \makecell{20.49\\\textcolor{gray}{\scriptsize$\pm$0.02}}  & \makecell{45.00\\\textcolor{gray}{\scriptsize$\pm$0.03}}  & \makecell{11.67\\\textcolor{gray}{\scriptsize$\pm$0.04}}  & \makecell{17.50\\\textcolor{gray}{\scriptsize$\pm$0.06}}  & \makecell{13.33\\\textcolor{gray}{\scriptsize$\pm$0.02}}  & \makecell{10.00\\\textcolor{gray}{\scriptsize$\pm$0.03}}  & \makecell{31.67\\\textcolor{gray}{\scriptsize$\pm$0.06}}  & \makecell{23.69\\\textcolor{gray}{\scriptsize$\pm$0.01}}  \\
\includegraphics[width=0.4cm]{icons/openai.png} GPT-4 \cite{openai2023gpt4} & N/A & \makecell{26.30\\\textcolor{gray}{\scriptsize$\pm$0.02}}  & \makecell{71.25\\\textcolor{gray}{\scriptsize$\pm$0.03}}  & \makecell{48.33\\\textcolor{gray}{\scriptsize$\pm$0.06}}  & \makecell{40.00\\\textcolor{gray}{\scriptsize$\pm$0.08}}  & \makecell{27.50\\\textcolor{gray}{\scriptsize$\pm$0.03}}  & \makecell{15.00\\\textcolor{gray}{\scriptsize$\pm$0.04}}  & \makecell{28.33\\\textcolor{gray}{\scriptsize$\pm$0.06}}  & \makecell{37.92\\\textcolor{gray}{\scriptsize$\pm$0.01}}  \\
        \midrule
        \rowcolor[RGB]{234, 238, 234}
        \multicolumn{10}{c}{\bf One-shot CoT Prompting} \\
        \midrule
\includegraphics[width=0.4cm]{icons/pmc.png} PMC-LLaMA \cite{wu2024pmc} & 13B & \makecell{5.20\\\textcolor{gray}{\scriptsize$\pm$0.01}}  & \makecell{10.42\\\textcolor{gray}{\scriptsize$\pm$0.02}}  & \makecell{8.33\\\textcolor{gray}{\scriptsize$\pm$0.04}}  & \makecell{2.50\\\textcolor{gray}{\scriptsize$\pm$0.02}}  & \makecell{7.08\\\textcolor{gray}{\scriptsize$\pm$0.02}}  & \makecell{1.25\\\textcolor{gray}{\scriptsize$\pm$0.01}}  & \makecell{11.67\\\textcolor{gray}{\scriptsize$\pm$0.04}}  & \makecell{6.97\\\textcolor{gray}{\scriptsize$\pm$0.01}}  \\
\includegraphics[width=0.4cm]{icons/meditron.png} \textsc{MediTron} \cite{chen2023meditron} & 70B & \makecell{22.94\\\textcolor{gray}{\scriptsize$\pm$0.02}}  & \makecell{39.58\\\textcolor{gray}{\scriptsize$\pm$0.03}}  & \makecell{31.67\\\textcolor{gray}{\scriptsize$\pm$0.06}}  & \makecell{15.00\\\textcolor{gray}{\scriptsize$\pm$0.06}}  & \makecell{20.42\\\textcolor{gray}{\scriptsize$\pm$0.03}}  & \makecell{15.00\\\textcolor{gray}{\scriptsize$\pm$0.04}}  & \makecell{31.67\\\textcolor{gray}{\scriptsize$\pm$0.06}}  & \makecell{26.27\\\textcolor{gray}{\scriptsize$\pm$0.01}}  \\
\includegraphics[width=0.4cm]{icons/mistral.png} Mistral \cite{jiang2023mistral} & 7B & \makecell{11.01\\\textcolor{gray}{\scriptsize$\pm$0.02}}  & \makecell{30.42\\\textcolor{gray}{\scriptsize$\pm$0.03}}  & \makecell{6.67\\\textcolor{gray}{\scriptsize$\pm$0.03}}  & \makecell{0.00\\\textcolor{gray}{\scriptsize$\pm$0.00}}  & \makecell{16.25\\\textcolor{gray}{\scriptsize$\pm$0.02}}  & \makecell{6.25\\\textcolor{gray}{\scriptsize$\pm$0.03}}  & \makecell{18.33\\\textcolor{gray}{\scriptsize$\pm$0.05}}  & \makecell{16.05\\\textcolor{gray}{\scriptsize$\pm$0.01}}  \\
\includegraphics[width=0.4cm]{icons/mistral.png} Mixtral \cite{jiang2024mixtral} & 8x7B & \makecell{28.13\\\textcolor{gray}{\scriptsize$\pm$0.02}}  & \makecell{50.83\\\textcolor{gray}{\scriptsize$\pm$0.03}}  & \makecell{8.33\\\textcolor{gray}{\scriptsize$\pm$0.04}}  & \makecell{22.50\\\textcolor{gray}{\scriptsize$\pm$0.07}}  & \makecell{21.25\\\textcolor{gray}{\scriptsize$\pm$0.03}}  & \makecell{8.75\\\textcolor{gray}{\scriptsize$\pm$0.03}}  & \makecell{33.33\\\textcolor{gray}{\scriptsize$\pm$0.06}}  & \makecell{29.23\\\textcolor{gray}{\scriptsize$\pm$0.01}}  \\
\includegraphics[width=0.4cm]{icons/meta.png} Llama 3 \cite{touvron2023llama2} & 8B & \makecell{34.86\\\textcolor{gray}{\scriptsize$\pm$0.03}}  & \makecell{35.42\\\textcolor{gray}{\scriptsize$\pm$0.03}}  & \makecell{3.33\\\textcolor{gray}{\scriptsize$\pm$0.02}}  & \makecell{2.50\\\textcolor{gray}{\scriptsize$\pm$0.02}}  & \makecell{20.00\\\textcolor{gray}{\scriptsize$\pm$0.03}}  & \makecell{11.25\\\textcolor{gray}{\scriptsize$\pm$0.04}}  & \makecell{41.67\\\textcolor{gray}{\scriptsize$\pm$0.06}}  & \makecell{27.13\\\textcolor{gray}{\scriptsize$\pm$0.01}}  \\
\includegraphics[width=0.4cm]{icons/meta.png} Llama 3 \cite{touvron2023llama2} & 70B & \makecell{41.59\\\textcolor{gray}{\scriptsize$\pm$0.03}}  & \makecell{56.25\\\textcolor{gray}{\scriptsize$\pm$0.03}}  & \makecell{30.00\\\textcolor{gray}{\scriptsize$\pm$0.06}}  & \makecell{22.50\\\textcolor{gray}{\scriptsize$\pm$0.07}}  & \makecell{27.50\\\textcolor{gray}{\scriptsize$\pm$0.03}}  & \makecell{27.50\\\textcolor{gray}{\scriptsize$\pm$0.05}}  & \makecell{45.00\\\textcolor{gray}{\scriptsize$\pm$0.06}}  & \makecell{39.45\\\textcolor{gray}{\scriptsize$\pm$0.02}}  \\
\includegraphics[width=0.4cm]{icons/openai.png} GPT-3.5 \cite{ouyang2022training} & N/A & \makecell{30.89\\\textcolor{gray}{\scriptsize$\pm$0.03}}  & \makecell{59.17\\\textcolor{gray}{\scriptsize$\pm$0.03}}  & \makecell{41.67\\\textcolor{gray}{\scriptsize$\pm$0.06}}  & \makecell{15.00\\\textcolor{gray}{\scriptsize$\pm$0.06}}  & \makecell{23.33\\\textcolor{gray}{\scriptsize$\pm$0.03}}  & \makecell{17.50\\\textcolor{gray}{\scriptsize$\pm$0.04}}  & \makecell{35.00\\\textcolor{gray}{\scriptsize$\pm$0.06}}  & \makecell{34.86\\\textcolor{gray}{\scriptsize$\pm$0.01}}  \\
\includegraphics[width=0.4cm]{icons/openai.png} GPT-4 \cite{openai2023gpt4} & N/A & \makecell{51.68\\\textcolor{gray}{\scriptsize$\pm$0.03}}  & \makecell{77.50\\\textcolor{gray}{\scriptsize$\pm$0.03}}  & \makecell{46.67\\\textcolor{gray}{\scriptsize$\pm$0.06}}  & \makecell{37.50\\\textcolor{gray}{\scriptsize$\pm$0.08}}  & \makecell{33.75\\\textcolor{gray}{\scriptsize$\pm$0.03}}  & \makecell{27.50\\\textcolor{gray}{\scriptsize$\pm$0.05}}  & \makecell{53.33\\\textcolor{gray}{\scriptsize$\pm$0.06}}  & \makecell{50.91\\\textcolor{gray}{\scriptsize$\pm$0.02}}  \\
        \bottomrule
    \end{tabular}
    \label{tab:main_results}
\end{table}

\subsection{Main Results}

Table \ref{tab:main_results} presents our evaluation results of various LLMs on the 1047 instances from \textsc{MedCalc-Bench}. From the table, we can observe the diverse performance of the models in different settings. In general, LLMs tend to perform better with the help of CoT prompting and one-shot learning, as evidenced by the improved accuracy for each LLM shown in the table. Among all LLMs compared, GPT-4 achieves the best performance in all three settings. In the zero-shot direct promoting setting, GPT-4 has a mean accuracy of 20.82\% on the task. By leveraging its own reasoning ability, the performance of GPT-4 can be improved to 37.92\%. Incorporating external medical knowledge from a one-shot demonstration further increases its accuracy to 50.91\%. Similar patterns can also be observed in many other LLMs, such as LLama 3 and Mixtral.

In addition to the general trend across different settings, the table also shows how various types of LLMs perform differently on our \textsc{MedCalc-Bench} test. While GPT-4 performs the best in our evaluation, the open-source Llama 3-70B model shows a competitive performance that is close to GPT-4. In both zero-shot direct prompting and zero-shot CoT prompting settings, Llama 3-70B achieves mean accuracies that are comparable to the results of GPT-4. However, GPT-4 significantly outperforms LLama 3-70B with the one-shot demonstration, which reflects its superior in-context learning capability for medical calculation. Moreover, by comparing Llama 3-8B/Mistral-7B with Llama 3-70B/Mixtral-8x7B, we find larger LLMs generally perform better on the medical calculation tasks, corresponding to the empirical scaling laws \cite{hoffmann2022training,kaplan2020scaling}. Interestingly, the 70B \textsc{MediTron} cannot beat Mistral-7B in the zero-shot settings, which can be explained by its poor instruction-following capability as the officially released model has not been instruction-tuned. With the additional demonstration in a one-shot setting, \textsc{MediTron} effectively learns the task and shows an improved performance close to Mixtral-8x7B.

It can also be observed from the table that the results for different subtasks in \textsc{MedCalc-Bench} present distinct patterns. For example, the performance of GPT-4 on the physical value calculation task is improved by 36.67\% by adding CoT prompting, while including an additional demonstration only further increases its accuracy by 6.25\%. In contrast, GPT-4 performance is improved by 8.33\% and 25\%, respectively, on the diagnosis calculation task, with the help of CoT prompting and one-shot demonstration. This result reflects that GPT-4 already contains certain medical knowledge concerning physical value calculation, thus the CoT prompting alone can significantly enhance its performance on such tasks. Nevertheless, the diagnosis calculation information is insufficient in its parametric knowledge, so the extra one-shot demonstration offers better help compared to the CoT prompting. Such an analysis enables us to have insights into the capabilities and limitations of LLMs on various medical calculation tasks, suggesting their different use cases in real-world applications.

\section{Discussion}

In this section, we provide an in-depth analysis of errors made by LLMs on \textsc{MedCalc-Bench}.

\subsection{What types of errors can LLMs make in \textsc{MedCalc-Bench}?}

We categorize four types of errors that LLMs can make in our dataset: 
\textbf{Type A (knowledge errors):} the model does not have the correct knowledge of the equation or rules used in the medical calculation task; 
\textbf{Type B (extraction errors):} the model extracts the wrong parameters from the patient note;
\textbf{Type C (computation errors):} the model conducts the arithmetic incorrectly;
\textbf{Type D (other errors):} all other cases of errors. 
Specific examples of the first three error types are shown in Table~\ref{tab:error_type}.

It should be noted that these errors are not independent of each other. 
For example, LLMs usually recall the relevant calculator knowledge first, and then extract the relevant parameters, and finally conduct the computation.
If the model recalls the calculator incorrectly, it is highly likely that it cannot extract the correct set of relevant parameters. Hence, we only consider the earliest error if there are multiple error types (e.g., if the model has error types A and B, then the error type will be A).

\begin{table}[h] \small
    \centering
    \caption{Examples of three error types in \textsc{MedCalc-Bench}. The \sethlcolor{lightred}\hl{incorrect rationale}, \sethlcolor{yellow}\hl{\texttt{\# comments}}, and \sethlcolor{yellow}\hl{evidence} are highlighted.}
    \begin{tabular}{cp{2.1cm}p{4.7cm}p{4.7cm}}
        \toprule
        \textbf{Type} & \textbf{Question} & \textbf{Patient Note} & \textbf{Error of GPT-4 Zero-shot CoT} \\
        \midrule
        \bf A & What is the patient's Wells' criteria score for Deep Vein Thrombosis? & A 77-year-old man was diagnosed with RA in 2006 with positive anti-CCP antibodies and RF; MTX was used but was changed to GLM in 2016 (the recent dose interval was 6–7 weeks; the last dose was 5 weeks before vaccine administration) and maintained [...] & The Wells' criteria for Deep Vein Thrombosis (DVT) includes [...] the patient has a brain hemorrhage, which is more likely than a DVT. The patient scores 1 point for paralysis and \sethlcolor{lightred}\hl{1 point for an alternative diagnosis being less likely}. \sethlcolor{yellow}\hl{\texttt{\# Wrong point assignment}} [...] \\
        \midrule
        \bf B & What are the number of criteria met for the PERC Rule for Pulmonary Embolism (PE)? & A 61-year-old male presented to our emergency department complaining of [...] \sethlcolor{yellow}\hl{A computerized tomography with contrast of the chest revealed multiple filling defects within the pulmonary arterial system, indicative of bilateral acute on chronic thromboembolic disease.} [...] & The PERC Rule for Pulmonary Embolism (PE) includes eight criteria: [...] The patient is older than 50 years, his pulse rate is less than 100 bpm, his SaO2 is less than 94\%, \sethlcolor{lightred}\hl{there is no mention of hemoptysis, estrogen use, prior DVT or PE} \sethlcolor{yellow}\hl{\texttt{\# The patient had PE}}, recent surgery or trauma, or unilateral leg swelling. [...] \\
        \midrule
        \bf C & What is the patient's LDL cholestrol concentration? Please outuput your answer in terms of mg/dL. & A 37-year-old male presented with progressive swelling of feet and facial puffiness for 6 months. On physical examination, he had mild pallor, moderate pedal edema and cloudy cornea with a peripheral arcus in both eyes [...] serum haptoglobin of 64 mg/dl (normal range: 70–200 mg/dl), total cholesterol 215 mg/dl, [...] & The patient's  [...] LDL = Total cholesterol - HDL - (Triglycerides / 5). The values needed for this calculation are given in the note: Total cholesterol is 215 mg/dL, HDL is 10 mg/dL, and Triglycerides are 385 mg/dL. Answer: LDL = \sethlcolor{lightred}\hl{215 - 10 - (385 / 5) = 142} \sethlcolor{yellow}\hl{\texttt{\# Wrong computation}} mg/dL \\
        \bottomrule
    \end{tabular}
    \label{tab:error_type}
\end{table}

\subsection{What errors do different LLMs make?}

To analyze the errors made by different LLMs, we utilize GPT-4 to classify their error types by comparing the LLM output to the ground truth in \textsc{MedCalc-Bench}.
We manually evaluate the annotations of 200 randomly sampled explanation errors, and find the accuracy of GPT-4 error classifier to be 89\%.
As such, we apply it to analyze the mistakes of all CoT prompting results.

Table \ref{tab:error_analysis_results} shows the distribution of error types in different settings. 
Under the zero-shot CoT setting, most of the errors (more than 50\%) belong to Type A in all LLMs, suggesting that recalling the correct equations or rules for the corresponding medical calculation task is the biggest challenge when no exemplar is provided. While Type A error is dominant under the zero-shot setting, its error rate varies in different LLMs, e.g. 0.96 in PMC-LLaMA and 0.35 in GPT-4. Such a difference reflects the diverse levels of medical calculation knowledge acquired by various LLMs.

Unlike the distributions in the zero-shot setting, errors that occurred with the one-shot CoT prompting have less than 50\% being categorized as Type A, which is consistently observed in different LLMs. 
This shows the effectiveness of the one-shot exemplar in providing the background rule or equation needed for medical calculation.
With the decrease in Type A errors, more Type B and Type C errors are captured in the wrong answers. This reveals the deficiencies of current LLMs in attribute extraction and arithmetic computation, which are required capabilities to perform real-world medical calculations.

\begin{table}[h!] \small
    \centering
    \caption{Error type distribution of LLMs on \textsc{MedCalc-Bench}. Numbers in parentheses denote the relative proportions. Arrows indicate the proportion changes from zero-shot to one-shot learning.}
    \begin{tabular}{lccccc}
        \toprule
        \bf Model & \textbf{Type A Error} & \textbf{Type B Error} & \textbf{Type C Error} & \textbf{Type D Error} & \bf \textbf{Error Rate} \\
        \midrule
        \rowcolor[RGB]{234, 238, 234}
        \multicolumn{6}{l}{\it Zero-shot CoT Prompting} \\
PMC-LLaMA-13B & 0.96 (96\%) & 0.03 (3\%) & 0.00 (0\%) & 0.01 (1\%) & 1.00 \\
\textsc{MediTron}-70B & 0.97 (97\%) & 0.00 (0\%) & 0.02 (2\%) & 0.00 (0\%) & 1.00 \\
Mistral-7B & 0.72 (80\%) & 0.11 (12\%) & 0.06 (7\%) & 0.00 (0\%) & 0.89 \\
Mixtral-8x7B & 0.55 (71\%) & 0.11 (14\%) & 0.09 (11\%) & 0.02 (3\%) & 0.78 \\
Llama 3-8B & 0.60 (72\%) & 0.11 (13\%) & 0.13 (15\%) & 0.00 (0\%) & 0.84 \\
Llama 3-70B & 0.43 (67\%) & 0.10 (16\%) & 0.11 (17\%) & 0.00 (0\%) & 0.64 \\
GPT-3.5 & 0.38 (50\%) & 0.24 (31\%) & 0.13 (17\%) & 0.02 (2\%) & 0.76 \\
GPT-4 & 0.35 (57\%) & 0.19 (30\%) & 0.08 (13\%) & 0.00 (0\%) & 0.62 \\
        \midrule
        \rowcolor[RGB]{234, 238, 234}
        \multicolumn{6}{l}{\it One-shot CoT Prompting} \\
PMC-LLaMA-13B & 0.42 (46\%\(\downarrow\)) & 0.31 (34\%\(\uparrow\)) & 0.17 (19\%\(\uparrow\)) & 0.01 (1\%--) & 0.91 \\
\textsc{MediTron}-70B & 0.24 (33\%\(\downarrow\)) & 0.23 (32\%\(\uparrow\)) & 0.26 (35\%\(\uparrow\)) & 0.01 (1\%\(\uparrow\)) & 0.74 \\
Mistral-7B & 0.32 (38\%\(\downarrow\)) & 0.33 (40\%\(\uparrow\)) & 0.17 (20\%\(\uparrow\)) & 0.01 (1\%\(\uparrow\)) & 0.83 \\
Mixtral-8x7B & 0.27 (38\%\(\downarrow\)) & 0.23 (33\%\(\uparrow\)) & 0.19 (27\%\(\uparrow\)) & 0.01 (2\%\(\downarrow\)) & 0.71 \\
Llama 3-8B & 0.25 (34\%\(\downarrow\)) & 0.17 (24\%\(\uparrow\)) & 0.29 (40\%\(\uparrow\)) & 0.02 (2\%\(\uparrow\)) & 0.73 \\
Llama 3-70B & 0.20 (34\%\(\downarrow\)) & 0.12 (20\%\(\uparrow\)) & 0.23 (39\%\(\uparrow\)) & 0.05 (8\%\(\uparrow\)) & 0.60 \\
GPT-3.5 & 0.23 (36\%\(\downarrow\)) & 0.20 (30\%\(\downarrow\)) & 0.20 (31\%\(\uparrow\)) & 0.02 (2\%--) & 0.65 \\
GPT-4 & 0.20 (40\%\(\downarrow\)) & 0.13 (27\%\(\downarrow\)) & 0.16 (33\%\(\uparrow\)) & 0.00 (0\%--) & 0.49 \\
        \bottomrule
    \end{tabular}
    \label{tab:error_analysis_results}
\end{table}

\subsection{Limitations and future work}\label{sec:limitations}

While our study provides a first-of-its-kind dataset to evaluate the medical calculation capabilities of various LLMs, there are several main limitations that can be improved in future work: (1) Due to the difficulty of manual verification of each instance in \textsc{MedCalc-Bench}, our dataset is limited in size, containing only 1047 instances in total. (2) Sometimes, the entity needed for a calculator is mentioned multiple times in a patient note, making it hard to extract the correct one. While we specifically prompt the model to extract the value of an entity that the patient has on their first day of admission, this is occasionally difficult to determine, especially if the patient note describes multiple visits. (3) While we saw a significant improvement in model performance with the one-shot exemplar, benchmarking the model with few-shot instances may have further increased the accuracy. However, curating such patient notes for rule-based calculators would have been difficult, given the labor-intensiveness of having to synthesize patient notes often requiring many attributes.   

\section{Related Work}

\label{related_work}

\subsection{Language Model Evaluations in Medicine} 

Existing datasets for evaluating LLMs in biomedicine \cite{fries2022bigbio} have primarily focused on verbal reasoning through multiple choice questions such as PubMedQA \cite{jin2019pubmedqa}, MedQA \cite{jin2021disease}, MedMCQA \cite{pal2022medmcqa}, and the medical questions in MMLU \cite{hendrycks2020measuring}. However, these datasets are mainly focused on qualitative reasoning instead of quantitative computation. Additionally, the format of multi-choice questions does not reflect the actual clinical settings where a single answer or response must be determined without any options provided. In this work, we introduce \textsc{MedCalc-Bench}, the first dataset that measures the quantitative reasoning capabilities of LLMs in medicine in a realistic setting where the LLM must determine the answer by itself without the support of answer choices. 

\subsection{Language Model Evaluations in Mathematics}

Many efforts have been made to evaluate the mathematical and computation capability of LLMs in various settings. GSM8k \cite{cobbe2021training} and MATH \cite{hendrycks2021measuring} are two examples which focus on pure mathematical problems. 
However, these datasets with general settings may not reflect LLM performance in domain-specific applications.
While there exist mathematical and computation-oriented datasets for specific domains such as chemistry \cite{ouyang2023structured}, their lack of manually verified step-by-step explanations weakens their reliability for model evaluation. \textsc{MedCalc-Bench} not only serves as the first dataset for medical-focused calculations, but also provides explanations that are verified by human experts. 
A full comparison of various datasets can be found in Table \ref{tab:related_work}.

\begin{table}[h!]
    \centering
    \caption{Comparison of different datasets for LLM evaluation. Medical: tasks for medical evaluation; Knowledge: dataset tests knowledge to a particular domain; Qualitative (Qual) Reasoning: dataset tests qualitative reasoning; Comput.: dataset requires computation (i.e., quantitative reasoning); Non-MCQ: questions which have a single answer and without the use of multiple choices.}
    \begin{tabular}{lccccc}
        \toprule
         & Medical & Knowledge & Qual. Reasoning & Comput. & Non-MCQ \\
        \midrule
        MedQA \cite{jin2021disease} & \greencheck & \greencheck & \greencheck & \xmark & \xmark \\
        MedMCQA \cite{pal2022medmcqa} & \greencheck & \greencheck & \greencheck & \xmark & \xmark \\
        PubMedQA \cite{jin2019pubmedqa} & \greencheck & \xmark & \greencheck & \xmark & \xmark \\
        MMLU \cite{hendrycks2020measuring} & \greencheck & \greencheck & \greencheck & \xmark & \xmark \\
        GSM8k \cite{cobbe2021training} & \xmark & \xmark & \xmark & \greencheck & \greencheck \\
        MATH \cite{hendrycks2021measuring} & \xmark & \xmark & \xmark & \greencheck & \greencheck \\
        \textsc{MedCalc-Bench} & \greencheck & \greencheck & \greencheck & \greencheck & \greencheck \\
        \bottomrule
    \end{tabular}
    \label{tab:related_work}
\end{table}

\subsection{Tool Learning}

One of the key features of language agents is the capability to use tools \cite{qin2023tool, sumers2023cognitive, wang2024survey, zhuang2024toolqa}, such as code interpreters \cite{chen2023program, gao2023pal} and external APIs \cite{qin2023toolllm, schick2024toolformer}.
GeneGPT \cite{jin2024genegpt} and ChemCrow \cite{m2024augmenting} utilize domain functionalities for scientific discovery.
While OpenMedCalc \cite{goodell2023augmentation} and AgentMD \cite{jin2024agentmd} use medical calculators to augment LLMs, their evaluations are based on small-scale or automatically constructed datasets.
Our manually-reviewed \textsc{MedCalc-Bench} is much larger than their evaluation datasets and contains both natural language explanations as well as final numeric answers.

\section{Conclusion}

In conclusion, this study introduces \textsc{MedCalc-Bench}, the first dataset specifically designed to evaluate the capabilities of LLMs for medical calculations. 
Our evaluations show that while LLMs like GPT-4 exhibit potential, none are reliable enough for clinical use. 
The error analysis highlights areas for improvement, such as knowledge recall and computational accuracy. 
We hope our work serves as a call to further improve LLMs and make them more suitable for medical calculations.

\clearpage
\begin{ack}
This research was supported by the NIH Intramural Research Program, National Library of Medicine. Additionally, the contributions made by Soren Dunn were done using the Delta advanced computing and data resource which is supported by the National Science Foundation (award OAC tel:2005572) and the State of Illinois. Delta is a joint effort of the University of Illinois Urbana-Champaign (UIUC) and its National Center for Supercomputing Applications (NCSA).
\end{ack}

\section*{Ethics Statement}

\label{sec:ethics}
For curating the patient notes in \textsc{MedCalc-Bench}, we only use publicly available patient notes from published case report articles in PubMed Central and clinician-generated anonymous patient vignettes.
As such, \textbf{no identifiable personal health information is revealed in this study}.

While \textsc{MedCalc-Bench} is designed to evaluate the medical calculation capabilities of LLMs, it should be noted that the dataset is not intended for direct diagnostic use or medical decision-making without review and oversight by a clinical professional. 
Individuals should not change their health behavior solely on the basis of our study.

\section*{Broader Impacts}

\label{sec:impacts}
As described in Sec~\ref{intro}, medical calculators are commonly used in the clinical setting.
With the rapidly growing interest in using LLMs for domain-specific applications, healthcare practitioners might directly prompt chatbots like ChatGPT to perform medical calculation tasks.
However, the capabilities of LLMs in these tasks are currently unknown.
Since healthcare is a high-stakes domain and wrong medical calculations can lead to severe consequences, including misdiagnosis, inappropriate treatment plans, and potential harm to patients, it is crucial to thoroughly evaluate the performance of LLMs in medical calculations.
Surprisingly, the evaluation results on our \textsc{MedCalc-Bench} dataset show that all the studied LLMs struggle in the medical calculation tasks.
The most capable model GPT-4 achieves only 50\% accuracy with one-shot learning and chain-of-thought prompting.
As such, our study indicates that \textbf{current LLMs are not yet ready to be used for medical calculations}.

It should be noted that while high scores on \textsc{MedCalc-Bench} do not guarantee excellence in medical calculation tasks, failing in this dataset indicates that the models must not be considered for such purposes at all. 
In other words, we believe that passing \textsc{MedCalc-Bench} should be a necessary (but not sufficient) condition for a model to be used for medical calculation.

\bibliography{neurips_data_2024}
\bibliographystyle{abbrvnat}

\newpage
\appendix

\appendix
\startcontents[supplementary]
\printcontents[supplementary]{l}{1}
{\section*{\centering Supplementary Materials for {\textsc{MedCalc-Bench}}}}

\newpage

\section{Dataset Curation - Additional Details}

\subsection{Extraction Process}
\label{sec:additional_details}

As mentioned in Section 2, the patient notes used for \textsc{MedCalc-Bench} come from three main sources: (1) Publicly available patient notes from Open-Patients, (2) Notes synthesized from templates (A.2), (3) Notes synthesized from templates. 

The majority of instances of \textsc{MedCalc-Bench} comes from Open-Patients, a dataset of 180k publicly available patient notes which are aggregated from four different sources as shown in the table below:

\begin{longtable}{lccp{4.8cm}}
\caption{Publicly available patient notes which make up Open-Patients} \\
\toprule
\textbf{Source} & \textbf{\# Patients} & \textbf{Average Token Length} & \textbf{Note Type} \\
\midrule
\endfirsthead

TREC Clinical Decision Support & 90 & 105.1 & This track consists of datasets of 30 patient notes each for three separate years from 2014-2016. The motivation of this track was to challenge participants to obtain relevant articles that can help answer potential questions for a particular patient note. The patient notes 2014 and 2015 are synthetic patient notes hand-written by individuals with medical training, but the 2016 dataset consists of real patient summaries coming from electronic health records.\\
\midrule
TREC Clinical Trials & 125 & 137.7 & This track consists of 125 patient notes, where 50 notes are from the year of 2021 and 75 notes are from the year of 2022. This track was meant to have participants retrieve previous clinical trials from ClinicalTrials.gov that best match the symptoms described in the patient note. The notes from both tracks are synthetic notes written by individuals with medical training meant to simulate an admission statement from an electronic health record (EHR). \\
\midrule
MedQA-USMLE & 12,893 & 135.8 & Questions from a multiple-choice form a professional medical board exam which can include patient summaries and will ask questions about particular issues based on the summary. \\
\midrule
PMC-Patients & 167,034 & 484 & Patient summaries extracted from case reports from PubMedCentral \\
\bottomrule
\end{longtable}

\newpage
This collection is released on HuggingFace under the CC-BY-SA 4.0 license: \url{https://huggingface.co/datasets/ncbi/Open-Patients/tree/main}.

Using the Open-Patients dataset, we curated notes for each calculator using this three-stage pipeline:

\indent \textbf{1.} We used GPT-3.5-turbo to identify eligible patient notes for each calculator. We did this by shortlisting the notes that had at least one relevant required for a given calculator.

\indent \textbf{2.}  With the shortlisted notes for each calculator, we then checked whether the note met the eligibility criteria for using the calculator based on the requirements provided by MDCalc.

\indent \textbf{3.}  Lastly, using the clinically eligible notes for each calculator from the second step, we used GPT-4 to extract the remaining attributes. We kept the notes that had all of the numeric parameters needed for each calculator and which also had enough categorical variables inferred such that 50\% of the numeric and categorical attributes were present for a given patient note. 


The attribute extractions of these notes were then verified for by authors of this paper. 

\textbf{Extraction 1 Prompt:}

For more details on step 1, we provided a set of 32 parameters which cover at least one attribute needed for each of the 55 calculators. For each note in Open-Patients, we applied the prompt shown below to determine which of the 32 parameters could be extracted from each note. From this, we received a dictionary of the 32 parameter IDs and their extracted values for each note. 

\begin{tcolorbox}[colframe=blue!75!black, colback=blue!10!white, title=System Prompt]

You are a helpful assistant for extracting the values of medical parameters from a patient note. Here is the list of parameters to consider:

\begin{longtable}{| >{\raggedright\arraybackslash}p{2cm} | >{\raggedright\arraybackslash}p{3cm} | >{\raggedright\arraybackslash}p{2cm} | >{\raggedright\arraybackslash}p{4cm} |}
    \hline
    \textbf{Parameter ID} & \textbf{Parameter Name} & \textbf{Type} & \textbf{Values \& Units} \\
    \hline
    3 & weight & numerical & kg, lbs \\
    \hline
    4 & height & numerical & m, cm, in, ft \\
    \hline
    5 & creatinine & numerical & mg/dL, µmol/L \\
    \hline
    13 & Systolic Blood Pressure & numerical & mm Hg \\
    \hline
    16 & Albumin & numerical & g/dL, g/L \\
    \hline
    20 & Heart Rate or Pulse & numerical & beats per minute \\
    \hline
    28 & Total cholesterol & numerical & mg/dL, mmol/L \\
    \hline
    112 & Cerebrovascular disease history & categorical & True, False \\
    \hline
    133 & Absence of cough or coryza & categorical & True, False \\
    \hline
    60 & Initial troponin & categorical & less than or equal to normal limit, between the normal limit or up to three times the normal limit, greater than three times normal limit \\
    \hline
    61 & Aspartate aminotransferase & numerical & U/L \\
    \hline
    66 & Temperature & numerical & degrees fahrenheit, degrees celsius \\
    \hline
    73 & Sodium & numerical & mEq/L, mmol/L \\
    \hline
    91 & Glucose & numerical & mg/dL, mmol/L \\
    \hline
    93 & Blood Urea Nitrogen (BUN) & numerical & mg/dL, mmol/L \\
    \hline
    102 & respiratory rate & numerical & breaths per minute \\
    \hline
    106 & White blood cell count & numerical & count/µL, count/L, count/mm³ \\
    \hline
    6 & History of Congestive Heart Failure & categorical & True, False \\
    \hline
    31 & Bilirubin & numerical & mg/dL \\
    \hline
    63 & Platelet count & numerical & count (per billion)/L, count (per thousand)/µL \\
    \hline
    114 & Partial pressure of oxygen (PaO\(_2\)) & numerical & mm Hg \\
    \hline
    108 & FiO\(_2\) & numerical & \% \\
    \hline
    44 & Previously documented Deep Vein Thrombosis & categorical & True, False \\
    \hline
    8 & Stroke & categorical & True, False \\
    \hline
    9 & Transient Ischemic Attacks History & categorical & True, False \\
    \hline
    7 & Hypertension history & categorical & True, False \\
    \hline
    105 & Hematocrit & numerical & \% \\
    \hline
    107 & Glasgow Coma Score & numerical & nan \\
    \hline
    23 & Hemoptysis & categorical & True, False \\
    \hline
    32 & international normalized ratio & numerical & ratio \\
    \hline
    215 & Ideal Body Weight & numerical (IBW) & kg \\
    \hline
    47 & History of ischemic heart disease & categorical & True, False \\
    \hline
\end{longtable}

\end{tcolorbox}

\begin{tcolorbox}[colframe=blue!75!black, colback=blue!10!white, title=System Prompt (continued)]

Please check the parameters one-by-one, and output a JSON dict in the following format: Dict\{Str(Parameter\_ID): Str(value)\}. This JSON should report the parameter id and a value for the parameter in the patient note (if available) for all 32 parameters.
If there are multiple values for a given measurement or attribute, then please use the value recorded based on when the patient note was written. You should not be using values that the patient had post-treatment or values from a patient's history in the past.
For parameters whose 'Type' is 'categorical', the exact name provided in the 'Parameter Name' column might not be mentioned inside the patient note. For such categorical variables, you should do your best to infer the value taken by the parameter. Additionally, for parameters whose "Type" is categorical, please select from one of values separated by commas in the Values \& Units column. Examples:  \{"8": "True"\}, \{"164": "Uncomplicated"\}. If there is no mention of a categorical parameter by the exact name and it cannot easily be inferred, then the value of the parameter should be "Not Mentioned" as the value, e.g. \{"26": "Not Mentioned"\} or \{"90": "Not Mentioned"\}.

If the 'Type' of the parameter is 'numerical' and the parameter is mentioned, please extract the exact values and units from the patient note and separate them by spaces. Examples:  \{'4': 5 'ft' 4 'in'\}, \{'3': 41.2 kg\}, \{'13': 67 mm Hg\}, \{'20': 117 beats per minute\}. 
For numerical parameters whose label is a concentration (mass/volume), the mass units MUST be one of the following: [mol, mmol, µmol, pmol, g, mg, µg, kg, mEq] and the units of volume MUST be one of the following: ['L', 'dL', 'mL', 'µL', 'mm$^3$', 'cm$^3$', 'm$^3$']. Examples: \{'16': 15.5 g/L\}, \{'5': 2.34 mg/dL\}.
The only exception to this rule is for platelet count and white blood cell count whose mass unit is 'count' and the volume unit can be one of ['L', 'dL', 'mL', 'µL', 'mm$^3$', 'cm$^3$', 'm$^3$']. Examples: \{'106': 15034 count/µL\}, \{'63': 1750000 count/L]\}.
For all other numeric parameters whose labels are not in the form of concentrations (mass/volume), you MUST use one of the EXACT label names provided in the Values \& Type column. Examples: \{'66': 39 degrees celsius\}. Numerical attributes which are not mentioned in the patient note should have their value set to 'Not Mentioned.' 
Please output a JSON dict in the mentioned format, specifying the value for each parameter ID in the provided list.
\end{tcolorbox}

\begin{tcolorbox}[colframe=green!75!black, colback=green!20!white, title=User Prompt]

Your task is to extract the 32 medical attributes and their values provided to you from the following patient note:
\newline
\newline
A 23-year-old previously healthy male presented to the ED with complaints of a headache that was gradual in onset and had been present for the prior 24 hours. He noted some lightheadedness and dizziness while standing, which prompted him to present to the ED for evaluation. He was febrile to 100.5 degrees Fahrenheit (F) and tachycardic to 110 beats per minute (bpm). The remainder of his physical exam was grossly unremarkable with no meningeal signs or focal neurologic deficits. He was provided antipyretics and intravenous (IV) fluids with complete resolution of his symptoms and discharged home with a diagnosis of viral syndrome. Two days later, he returned to the ED with complaints of continued headache and fever. He recalled a dry, tickling throat which was brief and self-limited in the prior two days. He was tachycardic, but afebrile on exam. With the exception of his tachycardia, his physical exam was again unremarkable without an identifiable infectious source. Laboratory evaluation demonstrated a bandemia of 8\% (reference range 0) as well as mild transaminitis with alanine aminotransferase (ALT) 177 units per liter (U/L) (reference range 17) and aspartate aminotransferase (AST) 171 U/L (reference range 12\u201339). His rapid heterophile antibody test was positive. He was discharged home with precautions to avoid contact sports and to have repeated liver function tests performed by his primary care provider. Three days after his second ED visit, he returned with jaundice, dark urine, and with continued fever and fatigue. He denied sore throat, cough, chest pain, abdominal pain, vomiting, diarrhea, hematuria, dysuria, or rash. He was again febrile with a temperature of 100.9\u00b0 F and a pulse rate of 109 bpm. There was noticeable scleral icterus and diffuse jaundice. He was also noted to have multiple, palpable, posterior cervical lymph nodes.Laboratory evaluation was notable for a leukocytosis of 14.8 \u00d710$^3$ cells per microliter (mcL) (reference 4.0\u201310.5) with lymphocytic predominance of 24\% and thrombocytopenia of 99\u00d710$^3$ cells/mcL (reference range 150\u2013450). Comprehensive metabolic panel was notable for mild hyponatremia of 133 millimoles (mmol) per L (reference range 136\u2013145 mmol/L), total bilirubin of 7.93 milligrams per deciliter (mg/dL) (reference 0.15\u20131.00), direct bilirubin of 6.9 mg/dL (reference range <0.2\u20130.3), alkaline phosphatase of 198 U/L (reference range 40\u2013129), ALT of 753 U/L (reference range 17\u201363), and AST 692 U/L (reference range 12\u201339). Coagulation studies were within normal limits. Acetaminophen level was negative at <1.5 micrograms per milliliter (reference range 10\u201330). Hepatitis serologies were notable for a reactive hepatitis B virus core antibody, non-reactive hepatitis B core antibody IgM, positive hepatitis B surface antibody, and negative hepatitis B surface antigen consistent with immunity due to natural infection. Hepatitis C antibody was non-reactive. Human immunodeficiency virus testing was negative. Blood cultures were also negative. EBV heterophile antibodies were positive.A formal right upper quadrant ultrasound demonstrated a mildly enlarged liver with normal contour. The gallbladder was visualized and noted to be contracted. The gallbladder wall was noted to be mildly thickened with a measurement of 0.34 centimeters. There was no evidence of cholelithiasis. The patient was admitted to the hospital for supportive care and further laboratory evaluation. He was provided IV fluids, and liver function tests were trended every six hours. Liver enzymes gradually decreased and his jaundice resolved. His thrombocytopenia was thought to be related to acute hepatitis. Coagulation studies remained within normal limits. He was discharged from the hospital with a diagnosis of cholestatic hepatitis secondary to EBV. He followed up with internal medicine and had serial liver function tests over the subsequent weeks."
\newline
\newline
Remember to use the appropriate units for numerical parameters and to use one of the values mentioned in the Values \& Units column for categorical parameters. There should only be one extracted value for each parameter and never a list. For numerical parameters which are not explicitly mentioned, please set the value as 'Not Mentioned'. For categorical parameters whose value cannot be inferred, please set the value as 'Not Mentioned.'
You should output all of the parameter IDs and their values inside a JSON dict.

\end{tcolorbox}

\begin{tcolorbox}[colframe=red!75!black, colback=red!10!white, title=Output]
\begin{verbatim}
{
    "3": "Not Mentioned", 
    "4": "Not Mentioned", 
    "5": "Not Mentioned",
    "13": "110 mm Hg",
    "16": "Not Mentioned", 
    "20": "110 beats per minute", 
    "28": "Not Mentioned",
    "112": "Not Mentioned", 
    "133": "Not Mentioned", 
    "60": "Not Mentioned", 
    "61": "171 U/L",
    "66": "100.5 degrees Fahrenheit", 
    "73": "133 mmol/L", 
    "91": "Not Mentioned",
    "93": "Not Mentioned", 
    "102": "Not Mentioned", 
    "106": "14.8 x10^3 cells/mcL",
    "6": "Not Mentioned", 
    "31": "0.34 centimeters", 
    "63": "99 x10^3 cells/mcL",
    "114": "Not Mentioned", 
    "108": "Not Mentioned", 
    "44": "Not Mentioned",
    "8": "Not Mentioned", 
    "9": "Not Mentioned", 
    "7": "Not Mentioned",
    "105": "Not Mentioned", 
    "107": "Not Mentioned", 
    "23": "Not Mentioned",
    "32": "Not Mentioned", 
    "215": "Not Mentioned", 
    "47": "Not Mentioned"
}
\end{verbatim}
\end{tcolorbox}
\textbf{Step 2: Determine Clinical Eligibility}

 From step 1, we curated notes for each calculator that had at least one attribute needed for the calculator. We then determined whether the patient was clinically eligible based on the criteria provided by MDCalc.

Using GPT-3.5-turbo, we applied the following prompt to find the eligible notes for each calculator based on the shortlisted set of notes for each calculator in step 1. Shown below is an example for checking the eligibility for the CURB-65 calculator:

\begin{tcolorbox}[colframe=blue!75!black, colback=blue!10!white, title=System Prompt]

You are a helpful assistant that needs to determine if the patient meets the criteria for using the calculator, Curb-65 Score, based on a patient note the user provides. The eligibility criteria for using the following calculator is as follows:
\newline
\newline
You can have an explanation if you would like, but you must respond with ELIGIBILITY: YOUR\_RESPONSE\_HERE at the end of your response, where YOUR\_ANSWER\_HERE can either be Yes if the patient meets the criteria for using the calculator, or No if the patient does not meet the eligibility criteria for the calculator. It's ok if there are missing values for the calculator that are needed to compute the score for the patient, your only focus is to just check if the patient's condition is eligible or not to even use the calculator. If you cannot tell whether or not the patient's condition meets the criteria, do your best to make an inference.

\end{tcolorbox}

\begin{tcolorbox}[colframe=green!75!black, colback=green!20!white, title=User Prompt]
\textbf{User Prompt: }

Patient Note: A 70-year-old Caucasian male initially presented to the emergency department (ED) of our hospital with fever and chills, which began one hour prior to his presentation. He also reported nausea and a productive cough with greenish sputum. The patient had been previously admitted to our hospital for pneumonia a month prior to his presentation and was discharged to a rehabilitation facility for three weeks. His past medical history was significant for end stage renal disease (ESRD) with dialysis dependence, failed kidney transplant, coronary artery disease (CAD) status with four drug-eluting stents (DES), type 2 diabetes mellitus, chronic pneumonia, and hypertension. On admission, his vital signs were as follow: temperature 39.4°C, blood pressure 87/55 mm Hg, pulse 100 beats per minute (bpm), oxygen saturation 88\% on room air and respiratory rate 18 breaths per minute. Physical examination revealed coarse breath sounds bilaterally on auscultation, a left upper arm arteriovenous fistula as well as multiple superficial abrasions bilaterally on his lower extremities. His laboratory findings were as follows: white blood cells 15.12 K/µL, platelets 104 K/µL, sodium 134 mmol/L, blood urea nitrogen 39 mg/dl, creatinine 4.5 mg/dl, lactic acid 2.77 mmol/L and procalcitonin 25.24 ng/ml. Urine analysis was not performed because the patient was anuric. His chest X-ray showed evidence of small effusions and bibasilar airspace disease which were indicative of atelectasis or infection. An electrocardiogram showed normal sinus rhythm, a rate of 99 bpm, normal axis and no significant ST abnormalities, with paroxysmal ventricular contractions (PVCs). Blood cultures were drawn, and the patient received ceftriaxone and piperacillin/tazobactam empiric antibiotics and intravenous (IV) fluids. Gram stain of the blood specimen showed gram-negative rods.
He was initially admitted to the progressive care unit (PCU) with cardiac telemetry. He was found to have second-degree atrioventricular (AV) block and was subsequently transferred to the intensive care unit (ICU) for closer management. On the second day of admission, blood cultures grew Pasteurella multocida. The organism was sensitive to beta-lactams including ampicillin, amoxicillin, amoxicillin/clavulanic acid and ampicillin/sulbactam. The patient was treated with IV cefepime due to his comorbidities. At this time, the patient was further questioned about recent animal contact. He stated that he lived with his wife and 14 cats and admitted the cats usually scratched him when they laid on him. His subsequent hospital stay was unremarkable. He was discharged home six days after admission with amoxicillin/clavulanic acid for two weeks. The patient returned to the outpatient clinic two weeks after discharge and had made an uneventful recovery.
\newline
\newline
Does this patient meet the eligibility criteria for CURB-65 Score? Simply check based on whether the patient's condition meets the criteria and DO NOT base your answer based on whether the note mentions the required attributes needed for computing a score with that calculator. You can have an explanation if you would like, but at the end, your response MUST include either ELIGIBILITY: Yes or ELIGIBILITY: No somewhere inside your note.
\end{tcolorbox}

\begin{tcolorbox}[colframe=red!75!black, colback=red!10!white, title=Output]

ELIGIBILITY: Yes

Based on the information provided in the patient note, the patient meets the eligibility criteria for using the CURB-65 Score calculator. The calculator is used to risk stratify a patient's community-acquired pneumonia in the emergency department setting. Although some specific values required for computation of the CURB-65 Score are missing in the patient note, such as confusion, blood urea nitrogen level, respiratory rate, and age, the information provided in the note indicates that the patient has pneumonia symptoms with a productive cough, fever, and chills. Additionally, the patient has risk factors such as advanced age, comorbid conditions including end-stage renal disease, diabetes, and coronary artery disease, as well as abnormal vital signs including low blood pressure, tachycardia, and low oxygen saturation.

It is important to note that the CURB-65 Score is used to assess the severity of pneumonia and determine the need for hospitalization, but additional clinical judgment and evaluation are necessary for appropriate management and treatment decisions.
\end{tcolorbox}

\newpage
\textbf{Prompt for Step 3:}

Lastly, after curating the eligible notes for each calculator, we then extracted the remaining parameters needed for the calculator. We provide the following example of how the parameter extraction from a patient note is done for the CURB-65 Score calculator: 

\begin{tcolorbox}[colframe=blue!75!black, colback=blue!10!white, title=System Prompt]

You are a helpful assistant for extracting the values of medical parameters needed for CURB-65 Score for Pneumonia Severity from a patient note. Here is a table of parameters to consider:

\begin{longtable}{| >{\raggedright\arraybackslash}l | >{\raggedright\arraybackslash}p{2.5cm} | >{\raggedright\arraybackslash}l | >{\raggedright\arraybackslash}p{3cm} | >{\raggedright\arraybackslash}l |}
    \hline
    \textbf{Parameter ID} & \textbf{Parameter Name} & \textbf{Type} & \textbf{Values \& Units} & \textbf{Description} \\
    \hline
    102 & respiratory rate & numerical & breaths per minute & nan \\
    \hline
    13 & Systolic Blood Pressure & numerical & mm Hg & nan \\
    \hline
    14 & Diastolic Blood Pressure & numerical & mm Hg & nan \\
    \hline
    176 & Confusion & categorical & True, False & nan \\
    \hline
    2 & age & numerical & years, months, weeks, days & nan \\
    \hline
    93 & Blood Urea Nitrogen (BUN) & numerical & mg/dL, mmol/L & nan \\
    \hline
\end{longtable}

Please check the parameters one-by-one, and output a JSON dictionary in the following format: Dict\{Str(Parameter\_ID): Str(value)\}. This JSON should report the parameter id and a value for the parameter in the patient note (if available) for all the parameters mentioned.

If there are multiple values for a given measurement or attribute, then please use the value recorded based on when the patient note was written. You should not be using values that the patient had post-treatment or values from a patient's history in the past.

For parameters whose 'Type' is 'categorical', the exact name provided in the 'Parameter Name' column might not be mentioned inside the patient note. For such categorical variables, you should do your best to infer the value taken by the parameter. If available, you should use the information in the 'Description' column to help you get more context about a variable and it's values which may help with inferring the value. Additionally, for parameters whose "Type" is categorical, please select from one of values separated by commas in the Values \& Units column. Examples:  \{"8": "True"\}, \{"164": "Uncomplicated"\}. If there is no mention of a categorical parameter by the exact name and it cannot easily be inferred, then the value of the parameter should be "Not Mentioned" as the value, e.g. \{"26": "Not Mentioned"\} or \{"90": "Not Mentioned"\}. Numerical parameters should never be inferred if explicitly not given and should be taken to be  "Not Mentioned," if the parameter is not mentioned in the patient note.

If the 'Type' of the parameter is 'numerical' and the parameter is mentioned, please extract the exact values and units from the patient note and separate them by spaces. Examples:  \{'4': 5 'ft' 4 'in'\}, \{'3': 41.2 kg\}, \{'13': 67 mm Hg\}, \{'20': 117 beats per minute\}. 
For numerical parameters whose label is a concentration (mass/volume), the mass units MUST be one of the following: [mol, mmol, µmol, pmol, g, mg, µg, kg, mEq] and the units of volume MUST be one of the following: ['L', 'dL', 'mL', 'µL', 'mm$^3$', 'cm$^3$', 'm$^3$']. Examples: \{'16': 15.5 g/L\}, \{'5': 2.34 mg/dL\}.
The only exception to this rule is for platelet count and white blood cell count whose mass unit is 'count' and the volume unit can be one of ['L', 'dL', 'mL', 'µL', 'mm$^3$', 'cm$^3$', 'm$^3$']. Examples: \{'106': 15034 count/µL\}, \{'63': 1750000 count/L]\}.
For all other numeric parameters whose labels are not in the form of concentrations (mass/volume), you MUST use one of the EXACT label names provided in the Values \& Type column. Examples: \{'66': 39 degrees celsius\}.
Please output a JSON dict in the mentioned format, specifying the value for each parameter ID in the provided table.
\end{tcolorbox}

\begin{tcolorbox}[colframe=green!75!black, colback=green!20!white, title=User Prompt]

Your task is to extract the parameters used for that calculator from the following patient note:
\newline
\newline
A 70-year-old Caucasian male initially presented to the emergency department (ED) of our hospital with fever and chills, which began one hour prior to his presentation. He also reported nausea and a productive cough with greenish sputum. The patient had been previously admitted to our hospital for pneumonia a month prior to his presentation and was discharged to a rehabilitation facility for three weeks. His past medical history was significant for end stage renal disease (ESRD) with dialysis dependence, failed kidney transplant, coronary artery disease (CAD) status with four drug-eluting stents (DES), type 2 diabetes mellitus, chronic pneumonia, and hypertension.
On admission, his vital signs were as follow: temperature 39.4°C, blood pressure 87/55 mm Hg, pulse 100 beats per minute (bpm), oxygen saturation 88\% on room air and respiratory rate 18 breaths per minute. Physical examination revealed coarse breath sounds bilaterally on auscultation, a left upper arm arteriovenous fistula as well as multiple superficial abrasions bilaterally on his lower extremities. His laboratory findings were as follows: white blood cells 15.12 K/µL, platelets 104 K/µL, sodium 134 mmol/L, blood urea nitrogen 39 mg/dl, creatinine 4.5 mg/dl, lactic acid 2.77 mmol/L and procalcitonin 25.24 ng/ml. Urine analysis was not performed because the patient was anuric. His chest X-ray showed evidence of small effusions and bibasilar airspace disease which were indicative of atelectasis or infection. An electrocardiogram showed normal sinus rhythm, a rate of 99 bpm, normal axis and no significant ST abnormalities, with paroxysmal ventricular contractions (PVCs). Blood cultures were drawn, and the patient received ceftriaxone and piperacillin/tazobactam empiric antibiotics and intravenous (IV) fluids. Gram stain of the blood specimen showed gram-negative rods.
He was initially admitted to the progressive care unit (PCU) with cardiac telemetry. He was found to have second-degree atrioventricular (AV) block and was subsequently transferred to the intensive care unit (ICU) for closer management. On the second day of admission, blood cultures grew Pasteurella multocida. The organism was sensitive to beta-lactams including ampicillin, amoxicillin, amoxicillin/clavulanic acid and ampicillin/sulbactam. The patient was treated with IV cefepime due to his comorbidities. At this time, the patient was further questioned about recent animal contact. He stated that he lived with his wife and 14 cats and admitted the cats usually scratched him when they laid on him. His subsequent hospital stay was unremarkable. He was discharged home six days after admission with amoxicillin/clavulanic acid for two weeks. The patient returned to the outpatient clinic two weeks after discharge and had made an uneventful recovery.
\newline
\newline
Remember to use the appropriate units for numerical parameters and to use one of the values mentioned in the Values \& Units column for categorical parameters. For numerical parameters which are not explicitly mentioned, please set the value as 'Not Mentioned'. For categorical parameters which are not explicitly mentioned in the patient note, do your best to infer the value.
You should output all of the parameter IDs and their values inside a JSON dict.
\end{tcolorbox}

\begin{tcolorbox}[colframe=red!75!black, colback=red!10!white, title=Output]
\begin{verbatim}
{
    "2": "70 years",
    "13": "87 mm Hg",
    "14": "55 mm Hg",
    "93": "39 mg/dL",
    "102": "18 breaths per minute",
    "176": "Not Mentioned"
}
\end{verbatim}
\end{tcolorbox}

At the end of this step, we parsed the parameter extractions for each note to ensure that all numeric attributes had an integer or decimal value and categorical variables were either “Not Mentioned,” or exactly matched one of the options provided in the Values \& Units column. After manually verifying the parameter extractions, we then had 1047 instances which covered 34 calculators of our dataset. We capped a maximum of 20 notes for each calculator and so some calculators had less than 20 notes. The remaining 21 calculators had their notes and extracted parameters either produced using template-based functions implemented in Python or the notes and the needed parameters were handwritten by clinicians. 

\subsection{Templates for Synthetic Notes}

For the following 11 calculators, we did not acquire any patient notes from Open-Patients. Instead, we used a template implemented in Python to create a patient note with the necessary values needed for each of the calculators:

\begin{enumerate}
    \item QTc Calculators - \{Bazett, Framingham, Rautaharju, Hodges, Fredericia\}
    \item Target Body Weight
    \item MME Conversion
    \item Steroid Conversion
    \item Estimated Due Date
    \item Estimated Gestational Age
    \item Estimated Date of Conception
\end{enumerate}

For each of the notes, we take random, but clinically plausible values for the required attributes needed for a particular calculator using the \texttt{random} library from Python. Shown below is an example of how we generate a note using a template for the MME Conversion calculator:

\begin{lstlisting}[language=Python]
def mme_conversion():
    
    mme_drugs = ["Codeine", "FentaNYL buccal", "FentANYL patch", "HYDROcodone", "HYDROmorphone", "Methadone", "Morphine", "OxyCODONE", "OxyMORphone", "Tapentadol", "TraMADol"]
    
    drugs = random.sample(mme_drugs, 3)

    note = "The patient takes "

    input_parameters = {}

    for i in range(3):

        num_doses = random.randint(1, 3)
        num_amount = round(random.randint(1, 7)) * 10

        key_name_dose = drugs[i] + " Dose"
        key_name_dose_per_day = drugs[i] + " Dose Per Day"

        if drugs[i] == "FentaNYL buccal" or drugs[i] == "FentaNYL patch":
            input_parameters[key_name_dose] = [num_amount , "$\mu$g"]
        else:
            input_parameters[key_name_dose] = [num_amount , "mg"]
        
        input_parameters[key_name_dose_per_day] = [num_doses, "per day"]

        add_s = 's'

        if num_doses == 1:
            add_s = ''

        if i == len(drugs) - 1:
            note += f"and {num_amount} mg of {drugs[i]} {num_doses} time{add_s} a day."
        else:
            note += f"{num_amount} mg of {drugs[i]} {num_doses} time{add_s} a day, "

    return note, input_parameters
\end{lstlisting}

Hence, making calls to \texttt{mme\_conversion()} would generate patient notes which all follow the same structure, but with different drugs and dosage amounts. Here are three such instances generated by the \texttt{mme\_conversion()} function:

1. The patient takes 70 mg of OxyMORphone 3 times a day, 60 mg of Codeine 2 times a day, and 30 mg of FentaNYL buccal 1 time a day.

2. The patient takes 30 mg of TraMADol 2 times a day, 40 mg of OxyMORphone 1 time a day, and 50 mg of OxyCODONE 1 time a day.

3. The patient takes 60 mg of HYDROmorphone 1 time a day, 50 mg of Codeine 2 times a day, and 30 mg of Methadone 1 time a day.


We used a similar approach for the other 10 calculators for which we needed a synthesized template. These templates can be found in \texttt{synthesize\_patient\_note.py} file of the Github repository.

\subsection{Clinician Synthesized Notes}
For the following 10 rule-based calculators, there were no notes curated from PMC-Patients and so we had notes synthesized from clinicians based on a set of pre-annotated values: Revised Score for Cardiac Risk index, HAS-BLED Score, Charlson Comorbidity Index, PSI Score, Child-Pugh Score for Cirrhosis Mortality, Glasgow Coma Score, APACHE II, SODA, Caprini Score for Venous Thromboembolism.

For synthesizing the notes, the following instructions were provided:

\begin{tcolorbox}[colframe=blue!75!black, colback=blue!10!white, title=Synthetic Patient Note Instructions]
\begin{itemize}
    \item \textbf{Calculator Information:} For each calculator, we provide a link from MDCalc.com which computes the medical value associated with a calculator, given the necessary inputs. You can use this link to get more information about the calculator, the details of the attributes and the values needed for a calculator, and any details about how a computation is performed.
    \item \textbf{How to synthesize a patient note:} Once you familiarize yourself with the calculator, please take a look at the Excel sheet provided for a given calculator. Here is the information about each of the columns:
        \begin{itemize}
            \item The first column (“Variable Name”) lists all of the attribute names needed for a given calculator.
            \item The second column (“Values \& Units”) will correspond to the possible values or units that the attribute can take on. For attributes that are categorical and do not take an integer or decimal value, you \textbf{MUST} use one of the values provided specified in the “Values \& Units” column for that given attribute (i.e. for the attribute “Hepatic Disease History” if the “Values \& Units” column lists “True, False” as the possible attributes, then you must use one of these). If an attribute is numerical, then you MUST use an integer (i.e. 43 years for the attribute “age”) or a decimal value (i.e. 3.32 mg/dL for the attribute “Creatinine”).
            \item The third column (“Pre-Annotated Value”) will correspond to the assigned value for a given attribute for a specific patient note. These are the values that are initially assigned for each item, but you can change them if they are not at all humanly plausible. In this case, you should adhere to the rules mentioned in the second bullet point for assigning an appropriate value and units. As mentioned, you can skip some descriptive (non-numeric) attributes and in this case, the value should be “Not Mentioned.”
            \item For some calculators, there will be a fourth column (“Description”) that will provide more context about the attribute. It may also specify the criteria that need to be met if someone assigns a particular value for a given attribute.
        \end{itemize}
\end{itemize}

In all, you need to synthesize a patient note based on the attributes provided in the first column and assign these attributes to the values provided in the third column.

To make the patient note as authentic as possible, for attributes that take on categorical values, try to avoid using the same exact name of the value in your note.

For example, from the calculator, Charlson Comorbidity Index (CCI), one of the attributes checks for the presence of liver disease with the following values: (None, Mild, Moderate to Severe). Instead of saying “The patient’s liver disease severity is classified as ‘mild,’” write something like “The patient presents with symptoms suggestive of chronic hepatitis, including persistent fatigue, abdominal discomfort, and elevated liver enzymes, warranting further evaluation for viral hepatitis markers and liver function tests.” By not directly specifying the value associated with liver disease, the patient note reads more authentically and we can use this to benchmark an LLM’s ability to properly determine the severity of the patient’s liver disease when provided with the categories.

Similarly, for the attribute “Moderate to Severe CKD” in the CCI calculator, instead of writing, “the patient did not pass the screening for chronic kidney disease,” write something like “the patient’s serum creatinine is 1.23 mg/dL.” This is equally indicative that the “Moderate to Severe CKD” attribute in the CCI calculator should be false. For numerical attributes, you can directly specify the value, but you should try to vary the units associated with the value.

This is not to say that you can never directly specify an attribute with a value provided from the third column, but you should vary your methods of conveying the values for an attribute so that we can measure how good an LLM is at deducing the value for a given attribute.
\end{tcolorbox}

\begin{tcolorbox}[colframe=blue!75!black, colback=blue!10!white, title=Synthetic Patient Note Instructions (continued)]
For attributes whose values are categorical, it’s fine if you do not mention every attribute needed for a calculator. All numerical attributes and gender need to be mentioned (i.e. age, creatinine concentration, etc.). However, you must have up to 50\% of the attributes in total in your note and you should change attributes you are omitting for each of the 5 patient notes. In this case, for non-numeric attribute values that you do change, please update the third column value to be “Not Mentioned.” You should also add other medical values (both descriptive and numeric) that are not relevant to the calculator so that we can determine if an LLM can sift through the noise and extract the correct values.
\newline
\newline
Example for Wells’ Criteria for PE:

\begin{center}
    \begin{tabular}{|l|c|c|}
    \hline
    \textbf{Variable Name} & \textbf{Values \& Units} & \textbf{Pre-Annotated Value} \\
    \hline
    Clinical signs and symptoms of DVT & True, False & False \\
    \hline
    PE is \#1 diagnosis OR equally likely & True, False & True \\
    \hline
    Heart rate > 100 & beats per minute & 120 \\
    \hline
    Immobilization at least 3 days & True, False & True \\
    \hline
    Surgery in the previous 4 weeks & True, False & False \\
    \hline
    Previous, objectively diagnosed PE & True, False & False \\
    \hline
    Previous, objectively diagnosed DVT & True, False & False \\
    \hline
    Hemoptysis & True, False & True \\
    \hline
    Malignancy w/ treatment within 6 months & True, False & False \\
    \hline
    \end{tabular}
\end{center}

\textbf{Patient Note:}

A 52-year-old female presents with swelling and discomfort in her left calf for the past two days. She mentions a recent period of reduced activity due to being bedridden for three days after sustaining a minor injury. The patient reports experiencing shortness of breath and chest discomfort, which she attributes to a recent respiratory issue. She recalls a recent episode of mild coughing with a small amount of blood-tinged sputum two days ago. She denies any recent surgical procedures or major illnesses, apart from a health issue she had dealt with a few years back. On examination, there is tenderness and warmth noted over the left calf. Her heart rate is elevated at 120 beats per minute, blood pressure is 140/90 mmHg, and respiratory rate is 18 breaths per minute. She is afebrile with a temperature of 98.6°F. Given the combination of recent sickness, respiratory symptoms, and mild hemoptysis, further evaluation with imaging studies is planned to rule out any significant pathology. The patient's laboratory tests including complete blood count, electrolyte panel, and renal function tests are within normal limits.
\end{tcolorbox}

Based on these instructions, clinicians synthesized 20 notes for each of the 10 rule-based calculators. They also provided the extracted parameter values in the same dictionary format that was used for obtaining the ground truth parameter values from Open-Patients.

\subsection{Templates for Natural Language Explanation}
\label{sec:template_explanation}

A core component of \textsc{MedCalc-Bench} is the natural language explanations for showing how the final answer is obtained. For each of the 55 calculators, we implement a function that takes in the input parameters needed for the calculator and outputs a step-by-step natural language explanation of how the final answer is obtained.

It should be noted that the parameter extractions from Section \ref{sec:additional_details} needed to go through an additional parsing step before they could be passed into template-based explanation functions. Firstly, we mapped each extracted parameter ID to the input variable name that would be used in Python. Additionally, for numerical attributes such as creatinine concentration, age, weight, etc., we extracted the integer/decimal along with the label from the string. For categorical variables whose string value was either "True" or "False," we converted these strings to boolean literals. Lastly, for categorical variables that required a specific value based on the MDCalc list, we ensured the value given by GPT-4 matched one of the values provided by MDCalc.

Shown below is an example of this processing done for the input parameters for a patient note using the Glasgow-Blatchford Bleeding Score:

\textbf{Extracted Parameters:}
\begin{verbatim}
{
    "96": "False",
    "93": "34 mg/dL",
    "95": "False",
    "20": "80 beats per minute",
    "92": "13 g/dL",
    "1": "Female",
    "13": "90 mm Hg",
    "97": true,
    "94": true
}
\end{verbatim}

From this, we transform the extracted attribute names to their names as Python variables for the template-based explanation function. We also convert the values into a format that can be used for the template-based explanation functions:

\textbf{Python Input Parameters:}
\begin{verbatim}
{
    "hepatic_disease_history": false,
    "bun": [
        34.0,
        "mg/dL"
    ],
    "syncope": false,
    "heart_rate": [
        80.0,
        "beats per minute"
    ],
    "hemoglobin": [
        13.0,
        "g/dL"
    ],
    "sex": "Female",
    "sys_bp": [
        90.0,
        "mm Hg"
    ],
    "cardiac_failure": true,
    "melena_present": true
}
\end{verbatim}

It should be noted that not all categorical variables may be reported in the present in a patient note. Hence, if the value of the variable cannot be inferred, we state that it is not mentioned inside the template, and then report that we assume it to be false. 

Based on this structure for generating the inputs to the explanation-based functions, we have implemented a template for providing a step-by-step explanation for all 55 calculators. Here is an example for how Glasgow-Blatchford Bleeding Score (GBS) explanation template was implemented:

\begin{lstlisting}[language=Python, breaklines=true, basicstyle=\small\ttfamily]
def glasgow_bleeding_score_explanation(input_parameters):

    score = 0

    hemoglobin_exp, hemoglobin = unit_converter_new.conversion_explanation(input_parameters["hemoglobin"][0], "hemoglobin", 64500, None, input_parameters["hemoglobin"][1], "g/dL")
    bun_exp, bun = unit_converter_new.conversion_explanation(input_parameters["bun"][0], "BUN", 28.08, None, input_parameters["bun"][1], "mg/dL")
    gender = input_parameters["sex"]
    systiolic_bp = input_parameters["sys_bp"][0]
    heart_rate = input_parameters["heart_rate"][0]
    
    explanation = f"The current glasgow bleeding score is 0. The patient's gender is {gender}.\n"
    explanation += hemoglobin_exp 

    if gender == "Male":
        if 12 < hemoglobin <= 13:
            explanation += f"Because the patient is a male and the hemoglobin concentration is between 12 and 13 g/dL, we add one point, making the current score {score} + 1 = {score + 1}.\n"
            score += 1
        elif 10 <= hemoglobin < 12:
            explanation += f"Because the patient is a male and the hemoglobin concentration is between 10 and 12 g/dL, we add three points, making the current score {score} + 3 = {score + 3}.\n"
            score += 3
        elif hemoglobin < 10:
            explanation += f"Because the patient is a male and the hemoglobin concentration is less than 10 and 12 g/dL, we add six points, making the current score {score} + 6 = {score + 6}.\n"
            score += 6
        elif hemoglobin > 13:
            explanation += f"Because the patient is a male and the hemoglobin concentration is greater than 13 g/dL, we do not add any points, keeping the current score at {score}.\n"

    else:
        if 10 < hemoglobin <= 12:
            explanation += f"Because the patient is a female and the hemoglobin concentration is between 10 and 12 mg/dL, we add one point, making the current score {score} + 1 = {score + 1}.\n"
            score += 1
        elif hemoglobin < 10:
            explanation += f"Because the patient is a female and the hemoglobin concentration is less than 10 mg/dL, we add three points, making the current score {score} + 3 = {score + 3}.\n"
            score += 6
        elif hemoglobin > 12:
            explanation += f"Because the patient is a female and the hemoglobin concentration is greater than 12 mg/dL, we do not add any points, keeping the current score at {score}.\n"

    explanation += bun_exp

    if 18.2 <= bun < 22.4:
        explanation += f"The BUN concentration is between 18.2 and 22.4 mg/dL, and so we add two points, making the current score {score} + 2 = {score + 2}.\n"
        score += 2
    elif 22.4 <= bun < 28:
        explanation += f"The BUN concentration is between 22.4 and 28 mg/dL, and so we add three points, making the current score {score} + 3 = {score + 3}.\n"
        score += 3
    elif 28 <= bun < 70:
        explanation += f"The BUN concentration is between 28 and 70 mg/dL, and so we add four points, making the current score {score} + 4 = {score + 4}.\n"
        score += 4
    elif bun > 70:
        explanation += f"The BUN concentration is greater than 70 mg/dL, and so we add six points, making the current score {score} + 6 = {score + 6}.\n"
        score += 6
    elif bun < 18.2:
        explanation += f"The BUN concentration is less than 18.2 mg/dL, and so we do not make any changes to the score, keeping the score at {score}.\n"

    explanation += f"The patient's blood pressure is {systiolic_bp} mm Hg. "

    if 100 <= systiolic_bp < 110:
        explanation += f"Because the patient's systolic blood pressure is between 100 and 110 mm Hg, we increase the score by one point, making the current score {score} + 1 = {score + 1}.\n"
        score += 1
    elif 90 <= systiolic_bp < 100:
        explanation += f"Because the patient's systolic blood pressure is between 90 and 100 mm Hg, we increase the score by two points, making the current score {score} + 2 = {score + 2}.\n"
        score += 2
    elif systiolic_bp < 90:
        explanation += f"Because the patient's systolic blood pressure is less than 90 mm Hg, we increase the score by three points, making the current score {score} + 3 = {score + 3}.\n"
        score += 3
    elif systiolic_bp >= 110:
        explanation += f"Because the patient's systolic blood pressure is greater than or equal to 110 mm Hg, we do not add points to the score, keeping the current score at {score} + 3 = {score + 3}.\n"


    explanation += f"The patient's heart rate is {heart_rate} beats per minute. "

    if heart_rate >= 100:
        explanation += f"Because the heart rate is greater or equal to than 100 beats per minute, we increase the score by one point, making the current score {score} + 1 = {score + 1}.\n"
        score += 1
    else:
         explanation += f"Because the heart rate is less than 100 beats per minute, we do not change the score, keeping the current score at {score}.\n"


    default_parameters = {"melena_present": "melena", "syncope": "recent syncope", "hepatic_disease_history": "hepatic disease history", "cardiac_failure": "cardiac failure"}

    for parameter in default_parameters:
        if parameter not in input_parameters:
            explanation += f"The patient's status for {default_parameters[parameter]} is missing from the patient note and so we assume it is absent from the patient.\n"
            input_parameters[parameter] = False
            explanation += f"Hence, we do not add any points to the score, keeping it at {score}.\n"
        
        elif parameter in ['syncope', 'hepatic_disease_history', 'cardiac_failure'] and input_parameters[parameter]:
            explanation +=  f"The patient has a {default_parameters[parameter]}, and so we add two points to the current total, making the current total {score} + 2 =  {score + 2}.\n"
            score += 2
            
        elif input_parameters[parameter]:
            explanation +=  f"The patient has {default_parameters[parameter]} and so we add one point to the current total, making the current total {score} + 1 =  {score + 1}.\n"
            score += 1
       
        else:
            explanation +=  f"The patient's status for {default_parameters[parameter]} is reported to be absent for the patient, and so we do not add any points, keeping the current total at {score}.\n"
        
    explanation += f"The patient's Glasgow Bleeding Score is {score}.\n"

    return {"Explanation": explanation, "Answer": score, "Calculator Answer": glasgow_bleeding_score(input_parameters)}
\end{lstlisting}

After applying the GBS explanation function\texttt{glasgow\_bleeding\_score\_explanation}, onto the python input parameters provided in Section \ref{sec:template_explanation}, we get the following output explanation:

\texttt{The current Glasgow bleeding score is 0. The patient's gender is Female. The concentration of hemoglobin is 13.0 g/dL. Because the patient is a female and the hemoglobin concentration is greater than 12 mg/dL, we do not add any points, keeping the current score at 0.
The concentration of BUN is 34.0 mg/dL. The BUN concentration is between 28 and 70 mg/dL, and so we add four points, making the current score 0 + 4 = 4.The patient's blood pressure is 90.0 mm Hg. Because the patient's systolic blood pressure is between 90 and 100 mm Hg, we increase the score by two points, making the current score 4 + 2 = 6.The patient's heart rate is 80.0 beats per minute. Because the heart rate is less than 100 beats per minute, we do not change the score, keeping the current score at 6.The patient has melena and so we add one point to the current total, making the current total 6 + 1 = 7. The patient's status for recent syncope is reported to be absent for the patient, and so we do not add any points, keeping the current total at 7.The patient's status for hepatic disease history is reported to be absent for the patient, and so we do not add any points, keeping the current total at 7.The patient has a cardiac failure, and so we add two points to the current total, making the current total 7 + 2 = 9. The patient's Glasgow Bleeding Score is 9.}

\subsection{\textsc{MedCalc-Bench} Calculators Covered}

\textsc{MedCalc-Bench} involves 55 different calculators. Shown below is a table providing more information about each calculator:

\begin{longtable}{p{4.8 cm}llcc}
\caption{Information for each calculator in \textsc{MedCalc-Bench} test set} \\
\toprule
\textbf{Calculator Name} & \textbf{Type} & \textbf{Sub-Type} & \makecell{\textbf{Number of}\\\textbf{Attributes}} & \makecell{\textbf{Number of}\\\textbf{Notes}} \\
\midrule
\endfirsthead
\caption[]{Information for each calculator in \textsc{MedCalc-Bench} test dataset (continued)} \\
\toprule
\textbf{Calculator Name} & \textbf{Type} & \textbf{Sub-Type} & \makecell{\textbf{Number of}\\\textbf{Attributes}} & \makecell{\textbf{Number of}\\\textbf{Notes}} \\
\midrule
\endhead

\midrule
\multicolumn{5}{r}{\textit{Continued on next page}} \\
\endfoot

\bottomrule
\endlastfoot

Creatinine Clearance (Cockcroft-Gault Equation) & Equation-Based & Lab Test & 5 & 20 \\
CKD-EPI Equations for Glomerular Filtration Rate & Equation-Based & Lab Test & 3 & 20 \\
CHA2DS2-VASc Score for Atrial Fibrillation Stroke Risk & Rule-Based & Risk & 10 & 20 \\
Mean Arterial Pressure (MAP) & Equation-Based & Physical & 3 & 20 \\
Body Mass Index (BMI) & Equation-Based & Physical & 3 & 20 \\
Calcium Correction for Hypoalbuminemia & Equation-Based & Lab Test & 2 & 20 \\
Wells' Criteria for Pulmonary Embolism & Rule-Based & Risk & 9 & 20 \\
MDRD GFR Equation & Equation-Based & Lab Test & 4 & 20 \\
Ideal Body Weight & Equation-Based & Physical & 2 & 20 \\
QTc Bazett Calculator & Equation-Based & Physical & 2 & 20 \\
Child-Pugh Score for Cirrhosis Mortality & Rule-Based & Severity & 5 & 20 \\
Wells' Criteria for DVT & Rule-Based & Risk & 11 & 20 \\
Revised Cardiac Risk Index for Pre-Operative Risk & Rule-Based & Risk & 6 & 20 \\
HEART Score for Major Cardiac Events & Rule-Based & Risk & 13 & 20 \\
Fibrosis-4 (FIB-4) Index for Liver Fibrosis & Equation-Based & Lab Test & 4 & 20 \\
Centor Score (Modified/McIsaac) for Strep Pharyngitis & Rule-Based & Severity & 5 & 20 \\
Maintenance Fluids Calculations & Equation-Based & Physical & 1 & 20 \\
MELD Na (UNOS/OPTN) & Equation-Based & Lab Test & 6 & 5 \\
HAS-BLED Score for Major Bleeding Risk & Rule-Based & Risk & 10 & 20 \\
Sodium Correction for Hyperglycemia & Equation-Based & Lab Test & 2 & 20 \\
Glasgow-Blatchford Bleeding Score (GBS) & Rule-Based & Risk & 9 & 20 \\
Serum Osmolality & Equation-Based & Lab Test & 3 & 20 \\
HOMA-IR (Homeostatic Model Assessment for Insulin Resistance) & Equation-Based & Lab Test & 2 & 2 \\
Charlson Comorbidity Index (CCI) & Rule-Based & Risk & 18 & 20 \\
FeverPAIN Score for Strep Pharyngitis & Rule-Based & Diagnosis & 5 & 20 \\
Free Water Deficit & Equation-Based & Lab Test & 4 & 20 \\
Anion Gap & Equation-Based & Lab Test & 3 & 20 \\
Fractional Excretion of Sodium (FENa) & Equation-Based & Lab Test & 4 & 5 \\
LDL Calculated & Equation-Based & Lab Test & 3 & 20 \\
CURB-65 Score for Pneumonia Severity & Rule-Based & Risk & 6 & 20 \\
Framingham Risk Score for Hard Coronary Heart Disease & Equation-Based & Lab Test & 7 & 16 \\
PERC Rule for Pulmonary Embolism & Rule-Based & Diagnosis & 9 & 20 \\
SIRS Criteria & Rule-Based & Diagnosis & 5 & 20 \\
QTc Fridericia Calculator & Equation-Based & Physical & 2 & 20 \\
QTc Framingham Calculator & Equation-Based & Physical & 2 & 20 \\
QTc Hodges Calculator & Equation-Based & Physical & 2 & 20 \\
QTc Rautaharju Calculator & Equation-Based & Physical & 2 & 20 \\
Body Surface Area Calculator & Equation-Based & Physical & 2 & 20 \\
Adjusted Body Weight & Equation-Based & Physical & 3 & 20 \\
Delta Gap & Equation-Based & Lab Test & 3 & 20 \\
Delta Ratio & Equation-Based & Lab Test & 3 & 20 \\
Albumin Corrected Anion Gap & Equation-Based & Lab Test & 4 & 20 \\
Albumin Corrected Delta Gap & Equation-Based & Lab Test & 4 & 20 \\
Albumin Corrected Delta Ratio & Equation-Based & Lab Test & 4 & 20 \\
PSI Score: Pneumonia Severity Index for CAP & Rule-Based & Severity & 20 & 20 \\
Glasgow Coma Score (GCS) & Rule-Based & Severity & 3 & 20 \\
APACHE II Score & Rule-Based & Risk & 20 & 20 \\
Sequential Organ Failure Assessment (SOFA) Score & Rule-Based & Risk & 16 & 20 \\
Caprini Score for Venous Thromboembolism (2005) & Rule-Based & Risk & 31 & 20 \\
Estimated Due Date & Equation-Based & Date & 1 & 20 \\
Steroid Conversion Calculator & Equation-Based & Dosage & 11 & 20 \\
Target weight & Equation-Based & Physical & 2 & 20 \\
Morphine Milligram Equivalents (MME) Calculator & Equation-Based & Dosage & 14 & 20 \\
Estimated Date of Conception & Equation-Based & Date & 1 & 20 \\
Estimated Gestational Age & Equation-Based & Date & 2 & 20 \\
\end{longtable}

In addition to the 1,047 instances for \textsc{MedCalc-Bench}, we also curated a training dataset of 10,053 to fine-tune open-source LLMs. We followed the exact same procedure as the one we used for the testing dataset, except authors of this paper did not manually verify the parameter extractions from GPT-4. In spite of this, we still gained a significant increase in performance for the two open-source LLMs that we fine-tuned on using this dataset. More details on the training dataset can be found in Section \ref{sec:training_dataset}. 

\section{Dataset License and Usage}

\subsection{Dataset License}

As mentioned in Section \ref{sec:additional_details}, the notes come from Open-Patients, templates, or are handwritten by clinicians. The notes coming from the latter two are our property, but the notes from Open-Patients were created using existing datasets (TREC, MedQA, and PMC-Patients). We have verified that we can use these datasets for making \textsc{MedCalc-Bench}. 

Specifically, although there is no issued license for the TREC datasets, the clinical trials and clinical decision support datasets from TREC are government-released datasets that were released for public use and distribution. The MedQA data is released by the MIT License. Hence, we can use instances from this dataset for \textsc{MedCalc-Bench}. Lastly, the PMC-Patients dataset is released by the CC-BY-SA 4.0 license. This gives us permission to re-distribute the dataset for \textsc{MedCalc-Bench} as long as it is not for commercial use. Additionally, because the PMC-Patients dataset is released by CC-BY-SA 4.0 license, we had to release Open-Patients, and consequently \textsc{MedCalc-Bench}, by the same license. Hence, both the training and testing instances of the \textsc{MedCalc-Bench} dataset are released by the \textbf{CC-BY-SA 4.0 license} and can be distributed for any non-commercial purposes.

The dataset and all the code for reproducing our results can be found at: \texttt{https://github.com/ncbi-nlp/MedCalc-Bench}. Additionally, we provide access to the training and test sets for \textsc{MedCalc-Bench} in a Croissant format using HuggingFace dataset as well:  
\texttt{https://huggingface.co/datasets/ncbi/MedCalc-Bench}.

\subsection{Dataset Instance Metadata}

For the \textsc{MedCalc-Bench} dataset uploaded on HuggingFace, each instance in both the training and testing dataset of \textsc{MedCalc-Bench} is a row that contains the following information: 

\begin{itemize}
    \item \textbf{Row Number} - Specifies the index of the instance. 
    \item \textbf{Calculator ID} - Specifies the integer ID of the calculator. 
    \item \textbf{Calculator Name} - Specifies the name of the clinical calculation task. 
    \item \textbf{Category} - Specifies the sub-category of the calculator. For equation-based calculators, the options are lab test, dosage, date, or physical and for rule-based calculators, the options are risk, severity, and diagnosis. 
    \item \textbf{Output Type} - Specifies the format type that the calculator will return. The options are decimal, integer, date (MM/DD/YY), or time in terms of weeks and days (i.e. (17 weeks, 4 days)). 
    \item \textbf{Note ID} - Specifies the ID of the patient note. The ID of the note will either be the ID given by Open-Patients or it will be an integer value if the patient note was handwritten by clinicians or synthesized by a template. 
    \item \textbf{Patient Note} - Specifies the patient note which provides the information needed to compute the final answer.
    \item \textbf{Question} - Specifies the question that is asked to the model to compute a specific medical value based on a particular calculator. 
    \item \textbf{Relevant Entities} - Provides a dictionary of the parameters and their extracted values based on the patient note. 
    \item \textbf{Ground Truth Answer} - Specifies the ground truth value without any units for the medical value that needs to be calculated. 
    \item \textbf{Lower Limit} - For equation-based calculators whose output is a decimal, this value is 95\% of the ground truth answer value. For all other cases, the lower limit is the same as the ground-truth value. 
    \item \textbf{Upper Limit} -  For equation-based calculators whose output is a decimal, this is value is 105\% of the ground truth answer value. For all other cases, the upper limit is the same as the ground-truth value. 
    \item \textbf{Ground Truth Explanation} - A paragraph for the data instance providing a step-by-step explanation for how the final answer was obtained. 
\end {itemize}

\subsection{Reproducing Results}

The main results for our dataset are the performances of various LLMs under different prompt settings. These are all shown in Table 2. First, please create a conda environment and install packages in the \texttt{requirements.txt} file. Then, please add your OpenAI API key to this environment. To obtain these results, simply execute the following command: \texttt{python \texttt{run.py} -- model <model\_name> -- prompt <prompt\_style>}.

The 8 options for the \texttt{model} argument as follows:

\begin{itemize}
   \item Mistral 7B: mistralai/Mistral-7B-Instruct-v0.2
   \item Mixtral 8x7B: mistralai/Mixtral-8x7B-Instruct-v0.1
   \item Llama3-8B: meta-llama/Meta-Llama-3-8B-Instruct
   \item Llama3-70B: meta-llama/Meta-Llama-3-70B-Instruct
   \item Meditron-70B: epfl-llm/meditron-70b
   \item PMC-Llama-13B: axiong/PMC\_LLaMA\_13B
   \item GPT-3.5: OpenAI/gpt-3.5-turbo
   \item GPT-4: OpenAI/gpt-4
\end{itemize}

Additionally, we provide three options for the \texttt{prompt} argument:

\begin{itemize}
    \item Direct answer = \texttt{direct\_answer}
    \item Zero shot = \texttt{zero\_shot}
    \item One shot = \texttt{one\_shot\_cot}
\end{itemize}

All open-source LLMs are run on 4 A100-80B GPUs for model inference. We provide the settings for the first three prompts settings below:

\begin{table}[h!]
    \centering
    \caption{Number of tokens used for zero-shot direct prompting}
    \begin{tabular}{ccccccccc}
    \toprule
         PMC-LLaMA & \textsc{MediTron} & Mistral & Mixtral & Llama 3-8B & Llama 3-70B & GPT-3.5 & GPT-4 \\
     \midrule
         Input & 276.9k & 736.9k & 669.9k & 669.9k & 727.5k & 727.5k & 662.5k & 662.5k \\
        Output & 23 & 10.6k & 130.1k & 36.7k & 8.6k & 7.4k & 8.1k & 8.5k \\
    \bottomrule
    \end{tabular}
    \label{tab:tokens_direct}
\end{table}

\begin{table}[h!]
    \centering
    \caption{Number of tokens used for zero-shot CoT prompting}
    \begin{tabular}{ccccccccc}
    \toprule
         & PMC-LLaMA & \textsc{MediTron} & Mistral & Mixtral & Llama 3-8B & Llama 3-70B & GPT-3.5 & GPT-4 \\
    \midrule
        Input & 332.6k & 823.8k & 721.2k & 721.2k & 778.8k & 778.8k & 713.8k & 713.8k \\
        Output & 240.9k & 18.0k & 241.5k & 222.6k & 230.3k & 223.9k & 191.4k & 216.8k \\
    \bottomrule
    \end{tabular}
    \label{tab:tokens_zero}
\end{table}

\begin{table}[h!]
    \centering
    \caption{Number of tokens used for one-shot CoT prompting}
    \begin{tabular}{ccccccccc}
    \toprule
         & PMC-LLaMA & \textsc{MediTron} & Mistral & Mixtral & Llama 3-8B & Llama 3-70B & GPT-3.5 & GPT-4 \\
    \midrule
        Input & 861.3k & 1.4M & 1.8M & 1.8M & 1.8M & 1.8M & 1.8M & 1.8M \\
        Output & 222.7k & 387.7k & 263.9k & 290.3k & 360.4k & 351.7k & 327.7k & 330.4k \\
    \bottomrule
    \end{tabular}
    \label{tab:tokens_one}
\end{table}

Upon executing \texttt{run.py}, the results will be saved in a file called \texttt{{<model>}\_{<prompt>}.jsonl}. Each instance will have the following metadata associated with them:

\begin{verbatim}
{
  "Row Number": ,
  "Calculator Name": ,
  "Calculator ID": ,
  "Category": ,
  "Note ID": ,
  "Question": ,
  "LLM Answer": , 
  "LLM Explanation": ,
  "Ground Truth Answer": ,
  "Ground Truth Explanation": ,
  "Result": 
}
\end{verbatim}

Here is what each item means:

\begin{itemize}
    \item \textbf{“Row Number”} - specifies the row in the \textsc{MedCalc-Bench} CSV
    \item \textbf{“Calculator Name”} - specifies which calculator that is being covered by this instance
    \item \textbf{“Calculator ID”} - unique ID of the calculator
    \item \textbf{“Category”} - sub-category for the calculator
    \item \textbf{“Note ID”} - specifies the Note ID from \textsc{MedCalc-Bench}
    \item \textbf{“Patient Note”} - provides the patient note covered by this instance
    \item \textbf{“Question”} - question being asked by this instance
    \item \textbf{“LLM Answer”} - the final answer value given by the LLM
    \item \textbf{“LLM Explanation”} - the explanation provided by the LLM for the problem
    \item \textbf{“Ground Truth Answer”} - the ground truth answer for the instance
    \item \textbf{“Ground Truth Explanation”} - the ground truth explanation for the instance
    \item \textbf{“Result”} - either “Correct” or “Incorrect” by comparing the LLM Answer and Ground Truth Answer
\end{itemize}

Note that in the direct answer setting, the LLM is only expected to provide a direct answer and so the \textbf{“LLM Explanation”} section will be “N/A” for all of them.

\subsection{Author Statement}

Although we have verified that our dataset is available by CC-BY-SA 4.0 license, all the authors bear responsibility for any infringements. For any updates on the dataset that needs to be made (i.e. adding new calculators, more notes, ect), we will update the Github repository and still keep archives of previous versions of the dataset on both the repository and HuggingFace. 

\section{Training Dataset for Fine-Tuning on \textsc{MedCalc-Bench}}
\label{sec:training_dataset}
\subsection{Training Dataset}

In addition to the 1,047 manually verified instances for \textsc{MedCalc-Bench}, we also curated a training dataset of 10,053 instances consisting of patient notes, questions, final answers, and explanations using the same extraction process that we used for curating the test set. This training dataset contained notes for 40 calculators. These calculators either had over 20 patient notes from Open-Patients or were one of the 11 calculators which had their notes synthesized from a template-based function (80 notes were synthesized for these calculators). The remaining 15 calculators had no patient notes to train with inside the training dataset. Shown in the table below are the number of instances for each calculator in the training dataset, along with their calculator type and sub-type. 

\begin{longtable}{p{4.8cm}llcc}
\caption{Information about each calculator in \textsc{MedCalc-Bench} training dataset} \\
\toprule
\textbf{Calculator Name} & \textbf{Type} & \textbf{Sub-Type} & \makecell{\textbf{Number of}\\\textbf{Attributes}} & \makecell{\textbf{Number of}\\\textbf{Notes}} \\
\midrule
\endfirsthead

\caption[]{Information about each calculator in \textsc{MedCalc-Bench} training dataset (continued)} \\
\toprule
\textbf{Calculator Name} & \textbf{Type} & \textbf{Sub-Type} & \makecell{\textbf{Number of}\\\textbf{Attributes}} & \makecell{\textbf{Number of}\\\textbf{Notes}} \\
\midrule
\endhead

\midrule
\multicolumn{5}{r}{\textit{Continued on next page}} \\
\endfoot

\bottomrule
\endlastfoot

Creatinine Clearance (Cockcroft-Gault Equation) & Equation-Based & Lab Test & 5 & 157 \\
CKD-EPI Equations for Glomerular Filtration Rate & Equation-Based & Lab Test & 3 & 519 \\
CHA2DS2-VASc Score for Atrial Fibrillation Stroke Risk & Rule-Based & Risk & 10 & 517 \\
Mean Arterial Pressure (MAP) & Equation-Based & Physical & 3 & 948 \\
Body Mass Index (BMI) & Equation-Based & Physical & 3 & 519 \\
Calcium Correction for Hypoalbuminemia & Equation-Based & Lab Test & 2 & 212 \\
Wells' Criteria for Pulmonary Embolism & Rule-Based & Risk & 9 & 445 \\
MDRD GFR Equation & Equation-Based & Lab Test & 4 & 312 \\
Ideal Body Weight & Equation-Based & Physical & 2 & 621 \\
QTc Bazett Calculator & Equation-Based & Physical & 2 & 80 \\
Wells' Criteria for DVT & Rule-Based & Risk & 11 & 578 \\
HEART Score for Major Cardiac Events & Rule-Based & Risk & 13 & 222 \\
Fibrosis-4 (FIB-4) Index for Liver Fibrosis & Equation-Based & Lab Test & 4 & 120 \\
Maintenance Fluids Calculations & Equation-Based & Physical & 1 & 889 \\
Sodium Correction for Hyperglycemia & Equation-Based & Lab Test & 2 & 295 \\
Serum Osmolality & Equation-Based & Lab Test & 3 & 370 \\
FeverPAIN Score for Strep Pharyngitis & Rule-Based & Diagnosis & 5 & 42 \\
Free Water Deficit & Equation-Based & Lab Test & 4 & 296 \\
Anion Gap & Equation-Based & Lab Test & 3 & 216 \\
LDL Calculated & Equation-Based & Lab Test & 3 & 66 \\
CURB-65 Score for Pneumonia Severity & Rule-Based & Risk & 6 & 125 \\
PERC Rule for Pulmonary Embolism & Rule-Based & Diagnosis & 9 & 153 \\
SIRS Criteria & Rule-Based & Diagnosis & 5 & 252 \\
QTc Fridericia Calculator & Equation-Based & Physical & 2 & 80 \\
QTc Framingham Calculator & Equation-Based & Physical & 2 & 80 \\
QTc Hodges Calculator & Equation-Based & Physical & 2 & 80 \\
QTc Rautaharju Calculator & Equation-Based & Physical & 2 & 80 \\
Body Surface Area Calculator & Equation-Based & Physical & 2 & 929 \\
Target weight & Equation-Based & Physical & 2 & 80 \\
Adjusted Body Weight & Equation-Based & Physical & 3 & 570 \\
Delta Gap & Equation-Based & Lab Test & 3 & 331 \\
Delta Ratio & Equation-Based & Lab Test & 3 & 288 \\
Albumin Corrected Anion Gap & Equation-Based & Lab Test & 4 & 91 \\
Albumin Corrected Delta Gap & Equation-Based & Lab Test & 4 & 75 \\
Albumin Corrected Delta Ratio & Equation-Based & Lab Test & 4 & 74 \\
MME Conversion & Equation-Based & Dosage & 2 & 80 \\
Steroid Conversion & Equation-Based & Dosage & 2 & 80 \\
Estimated Due Date & Equation-Based & Date & 2 & 80 \\
Estimated Date of Conception & Equation-Based & Date & 2 & 80 \\
Estimated Gestational Age & Equation-Based & Date & 2 & 80 \\
\end{longtable}

\subsection{Training Details}

We fine-tuned Mistral-7B and Llama2-7B on our dataset. All fine-tuning runs are performed on a single 4xA100 40GB node. We use an adapted version of the Code-Act \cite{wang2024executable} training scripts which are based on a fork of Megatron-LLM \cite{epfmgtrn}. We train Llama2-7b and not Llama3-7B which was benchmarked in the main paper as this model is not currently supported by Megatron-LLM. Both training and test data are converted into a chatML form. Shown below is an example:

\textbf{System Message}

Below is a patient note as well as a medical question about the patient. 
Provide an accurate answer to the question based on the note. 
Explain your reasoning before stating your final answer and put your final answer 
at the end of your response in the format Answer: INSERT\_ANSWER

\textbf{User Message}

Patient Note:

A 78-year-old male with history of hypertension, dyslipidemia, and prosthetic aortic valve replacement presented to the emergency department with complaints of exertional shortness of breath for few days. Chest X-ray showed widened mediastinum. Subsequently, CT angiogram of the chest showed massively dilated ascending aorta with maximum diameter of 10.2 cm []. Transthoracic echocardiogram revealed normal ejection fraction with mild aortic insufficiency, well-functioning prosthetic aortic valve, and no aortic stenosis. His creatinine was 1.8 mg/dl, and he had received 150 ml of contrast for CT angiogram. It was necessary to evaluate his coronary arteries before surgery for AAA. There were two options in this case: coronary CT angiography or conventional coronary angiography. Coronary CT angiography would require about 150 ml of contrast.[] A diagnostic coronary angiogram can be done using about 50 ml of contrast.[] Given his renal insufficiency, we decided to do conventional coronary angiography. Since the ascending aorta and root were extremely dilated, we knew it would not be possible to engage the coronaries using regular catheters and standard technique. Hence, telescopic technique was used. Amplatz left 3 (AL3) 7F (French) guide catheter (90 cm) was initially used, and nonselective injection of contrast was done to see the coronary ostium [ and ]. After that, a 5F multipurpose (MP) catheter (110 cm) was telescoped through 7F AL3 guide catheter to engage the ostium of the left main coronary artery [ and ]. Using this technique, images of coronaries were obtained, and it showed minimal luminal irregularities in major epicardial coronary arteries. Total of 50 ml contrast was used. The patient underwent successful surgery with aortic valve replacement and excision of aneurysm with graft placement. The most likely etiology of giant aortic aneurysm in our patient was atherosclerosis. However, we did not have any record of his previous aortic valve surgery, and we do not know if it was a bicuspid aortic valve. In that case, he could have had aortopathy associated with bicuspid aortic valve.
    
\textbf{Question}: Using the MDRD GFR equation, what is the patient's Glomerular Filtration Rate (GFR) in terms of mL/min/1.73 m²?

\textbf{Response}:

The patient is 78 years old. The concentration of Creatinine is 1.8 mg/dL. The race of the patient is not provided, so the default value of the race coefficient is 1.0. The patient is male, so the gender coefficient is 1.0.
The patient's estimated GFR is calculated using the MDRD equation as:
The formula for GFR is given by: $175 \times \text{creatinine}^{-1.154} \times \text{age}^{-0.203} \times \text{race\_coefficient} \times \text{gender\_coefficient}$. The creatinine concentration is mg/dL. Plugging in these values will give us: $175 \times 1.8^{-1.154} \times 78^{-0.203} \times 1.0 \times 1.0 = 36.674$. Hence, the patient's GFR is 36.674 mL/min/1.73m².

Answer: 36.674

Following the implementation used for finetuning LLMs for CodeAct , we only calculate and optimize the loss on the assistant response, pack short instances into longer ones, and train using flash attention. We train for two epochs with a batch size of 32 on 4 parallel GPU's with a learning rate of 1e-5, 50 warmup steps, and cosine decay.

\subsection{Results}

From the above fine-tuning procedure, we achieved the following results on the \textsc{MedCalc-Bench} test set: 

\begin{table}[ht]
\centering
\caption{Results after fine-tuning open-source LLMs on \textsc{MedCalc-Bench} training dataset}
\begin{tabular}{lll}
\toprule
\textbf{Model} & \textbf{Original Accuracy} & \textbf{Fine-tuned Accuracy} \\ 
\midrule
Mistral-7B    & 10.79\%     & 49.19\%           \\ 
Llama-2-7B    & 1.53\%    & 45.75\%           \\ 
\bottomrule
\end{tabular}
\end{table}

Both fine-tuning results significantly increased their respective models’ performance with the lower Llama-2 performance likely due to its lower context length. Hence, we show the efficacy of the training dataset for \textsc{MedCalc-Bench} in improving a model's ability to recall the correct equations, retrieve relevant parameters, and correctly perform arithmetic. Although these results demonstrate a significant performance improvement that fine-tuning has on our dataset, this still shows that more work needs to be done in improving LLMs for them to be reliable clinical calculators.

\newpage

\section{Code-Augmented LLMs Prompt Setting}

\subsection{Code-Augmented LLM Performance}

To minimize a model's arithmetic errors on \textsc{MedCalc-Bench}, we instead prompt a model to write code for performing any arithmetic operations. The user will then execute the code and provide the output from the console. If there are any compiling issues, the model has up to 20 tries to output the answer. For further details on this implementation, please refer to the \texttt{generate\_code\_prompt.py} file in the repository. 

Due to limited compute, we only ran the code execution prompts for GPT-3.5 and GPT-4 and did not perform this experiment for open-source LLMs. Hence, we only included these results in the supplemental section and not the main paper as not all LLMs were benchmarked in this setting. 

Shown below are the results for zero-shot chain-of-thought (CoT) for GPT-3.5-turbo and GPT-4, along with the code-augmented results for GPT-3.5-turbo-16k and GPT-4:

\begin{table}[ht]
    \centering
    \caption{Comparison of GPT-3.5-turbo and GPT-4 in zero-shot CoT before and after adding code interpreter}
    \begin{tabularx}{\textwidth}{l>{\raggedright\arraybackslash}X>{\raggedright\arraybackslash}X>{\raggedright\arraybackslash}X>{\raggedright\arraybackslash}X}
    \toprule
    & \multicolumn{2}{c}{\textbf{GPT-3.5-turbo}} & \multicolumn{2}{c}{\textbf{GPT-4}} \\ 
    \cmidrule(lr){2-3} \cmidrule(lr){4-5}
    & \textbf{Zero-Shot CoT} & \textbf{Code Prompt} & \textbf{Zero-Shot CoT} & \textbf{Code Prompt}\\
    \midrule
    \textbf{Equation} &  &  &  &  \\ \midrule
    \quad Lab Test & 20.49\% & 30.04\% & 26.30\% & 49.85\% \\ 
    \quad Physical & 45.00\% & 56.76\% & 71.25\% & 81.66\% \\ 
    \quad Dosage & 17.50\% & 15.79\% & 40.00\% & 32.50\% \\ 
    \quad Date & 11.67\% & 28.33\% & 48.33\% & 43.33\% \\ \midrule
    \textbf{Rule} & &  &  &  \\ \midrule
    \quad Severity & 10.00\% & 13.33\% & 15.00\% & 18.75\% \\ 
    \quad Diagnosis & 31.67\% & 25.00\% & 28.33\% & 23.33\% \\ 
    \quad Risk & 13.33\% & 12.68\% & 27.50\% & 33.75\% \\ \midrule
    \textbf{Overall} & 23.69\% & 30.29\% & 37.92\% & 48.51\% \\ 
    \bottomrule
    \end{tabularx}
\end{table}

After augmenting with a code interpreter, the accuracies of GPT-3.5-turbo and GPT-4 have increased by 6.60\% and 10.59\%, respectively. Hence, we conclude that LLMs can significantly benefit from using a code interpreter to reduce its arithmetic mistakes and, thereby, become better at solving clinical calculations.

\newpage

\section{Additional Analysis}

\subsection{Accuracy vs. Number of Attributes}

\begin{figure}[ht]
    \centering
    \begin{minipage}[b]{0.75\textwidth}
        \centering
        \includegraphics[width=\textwidth]{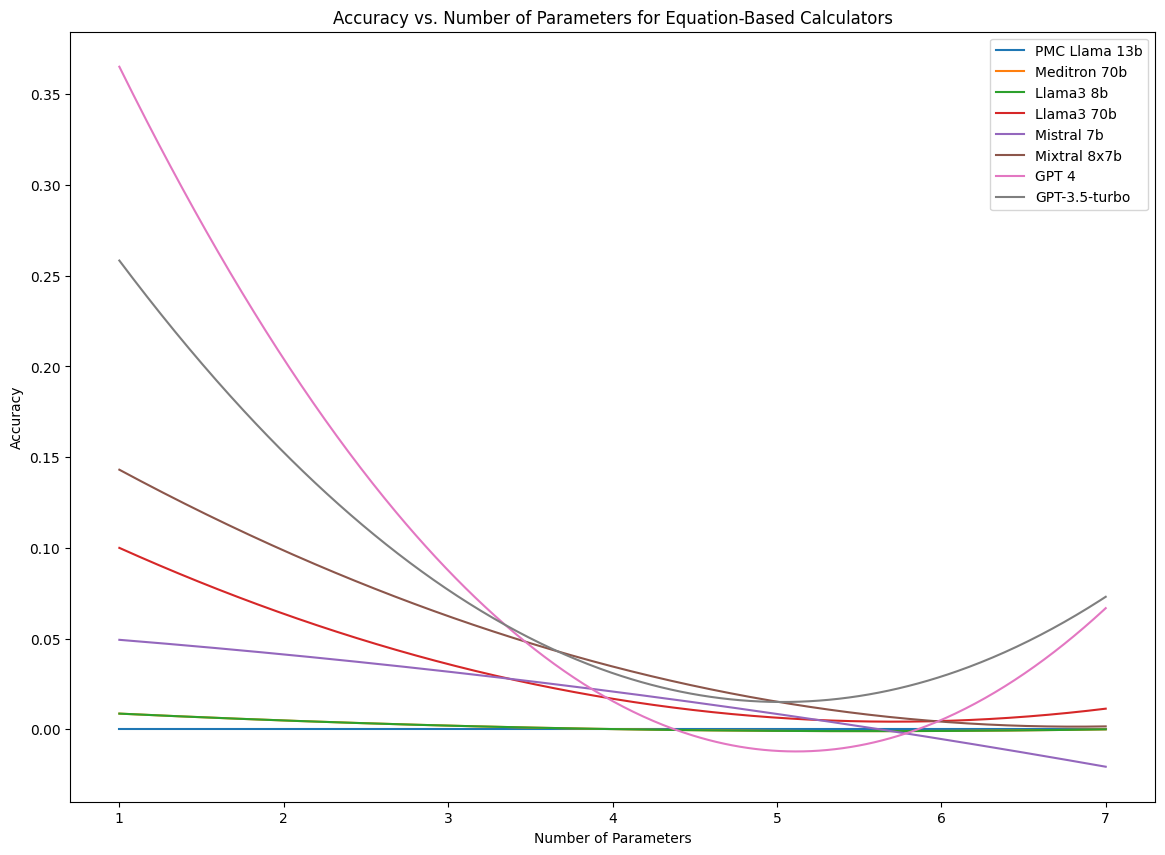}
        \caption{Accuracy vs. Number of Parameters for Equation-Based Calculators}
        \label{fig:plot1}
    \end{minipage}
    \hfill
    \hfill
    \begin{minipage}[b]{0.75\textwidth}
        \centering
        \includegraphics[width=\textwidth]{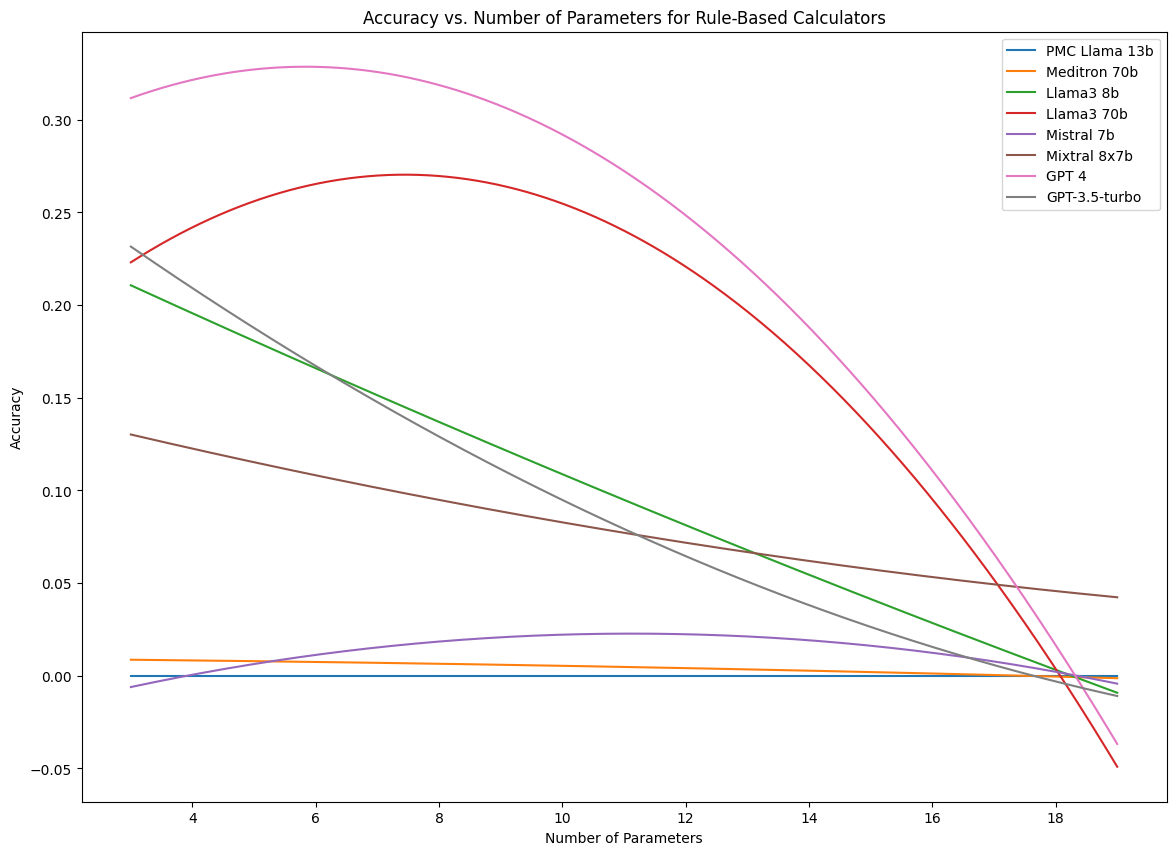}
        \caption{Accuracy vs. Number of Parameters for Rule-Based Calculators}
        \label{fig:plot2}
    \end{minipage}
\end{figure}

While we have conducted analyses for LLM performance based on sub-category type, we also examine the performance for models based on the number of parameters. Using a line-of-best-fit for a second degree polynomial, we see the following trend for the number of parameters vs. accuracy for each of the eight models. Figure \ref{fig:plot1} shows the results for equation-based calculators and Figure \ref{fig:plot2} shows the results for rule-based calculators. As seen, the overall trend is that model performance tends to drop as the number of attributes increases. For models such as Meditron-70B and PMC-Llama-13B the drop is not as drastic, but that is because the accuracy was already close to 0 even for calculation tasks with the minimum number of attributes. However, for models such as Llama3-70B and GPT-4 which perform well relative to the other LLMs, the drop is much more drastic as the parameter count increases. Hence, these plots show that even the highest-performing LLMs are at best able to make progress only on basic computation tasks and struggle significantly as the number of attributes increase.

\end{document}